\documentclass[10pt,twocolumn,letterpaper]{article}

\usepackage[pagenumbers]{wacv} %

\makeatletter
\@namedef{ver@everyshi.sty}{}
\makeatother
\usepackage{tikz}

\usepackage{graphicx}
\usepackage{amsmath}
\usepackage{amssymb}
\usepackage{booktabs}
\usepackage{comment}
\usepackage{colortbl}
\usepackage{multirow}
\usepackage{makecell}
\newif\ifarxiv
\arxivtrue

\def\MYTITLE{Combined Physics and Event Camera Simulator for Slip Detection}

\usepackage[breaklinks,colorlinks,bookmarks=true,bookmarksnumbered=true]{hyperref}
\hypersetup{
  pdftitle={\MYTITLE},
  pdfsubject={Robotics, Computer Vision, Artificial Intelligence},
  pdfauthor={Thilo Reinold, Suman Ghosh, Guillermo Gallego},
  pdfkeywords={Event Cameras, Slip Detection, Machine-Learning, Intelligent Systems}
}

\usepackage[capitalize]{cleveref}
\crefname{section}{Sec.}{Secs.}
\Crefname{section}{Section}{Sections}
\Crefname{table}{Table}{Tables}
\crefname{table}{Tab.}{Tabs.}

\def\wacvPaperID{3} %
\def\confName{WACV}
\def\confYear{2025}

\usepackage[absolute]{textpos}

\begin{document}

\ifarxiv
\definecolor{somegray}{gray}{0.5}
\newcommand{\darkgrayed}[1]{\textcolor{somegray}{#1}}
\begin{textblock}{11}(2.5, 0.8)  %
\begin{center}
\darkgrayed{This paper has been accepted for publication at the IEEE/CVF Winter Conference on\\ Applications of Computer Vision (WACV) Workshops, Tucson (USA), 2025.
\copyright IEEE}
\end{center}
\end{textblock}
\fi

\title{\MYTITLE}

\author{Thilo Reinold$^{1}$, Suman Ghosh$^{1}$ and Guillermo Gallego$^{1,2}$.\\
$^{1}$~Technische Universit\"at Berlin, and~Robotics Institute Germany.\\
$^{2}$~Einstein Center for Digital Future, and~Science of Intelligence Excellence Cluster.}
\maketitle

\begin{abstract}
Robot manipulation is a common task in fields like industrial manufacturing. 
Detecting when objects slip from a robot's grasp is crucial for safe and reliable operation. 
Event cameras, which register pixel-level brightness changes at high temporal resolution (called ``events''), offer an elegant feature when mounted on a robot's end effector: since they only detect motion relative to their viewpoint, a properly grasped object produces no events, while a slipping object immediately triggers them.
To research this feature, representative datasets are essential, both for analytic approaches and for training machine learning models. 
The majority of current research on slip detection with event-based data is done on real-world scenarios and manual data collection, as well as additional setups for data labeling. 
This can result in a significant increase in the time required for data collection, a lack of flexibility in scene setups, and a high level of complexity in the repetition of experiments.
This paper presents a simulation pipeline for generating slip data using the described camera-gripper configuration in a robot arm, and demonstrates its effectiveness through initial data-driven experiments. 
The use of a simulator, once it is set up, has the potential to reduce the time spent on data collection, provide the ability to alter the setup at any time, simplify the process of repetition and the generation of arbitrarily large data sets.
Two distinct datasets were created and validated through visual inspection and artificial neural networks (ANNs). 
Visual inspection confirmed photorealistic frame generation and accurate slip modeling,
while three ANNs trained on this data achieved high validation accuracy and demonstrated good generalization capabilities on a separate test set, along with initial applicability to real-world data.\\
Project page: \url{https://github.com/tub-rip/event_slip}
\end{abstract}

\section{Introduction}
\label{sec:intro}
\begin{figure}[t]
  \centering
    \includegraphics[width=\linewidth]{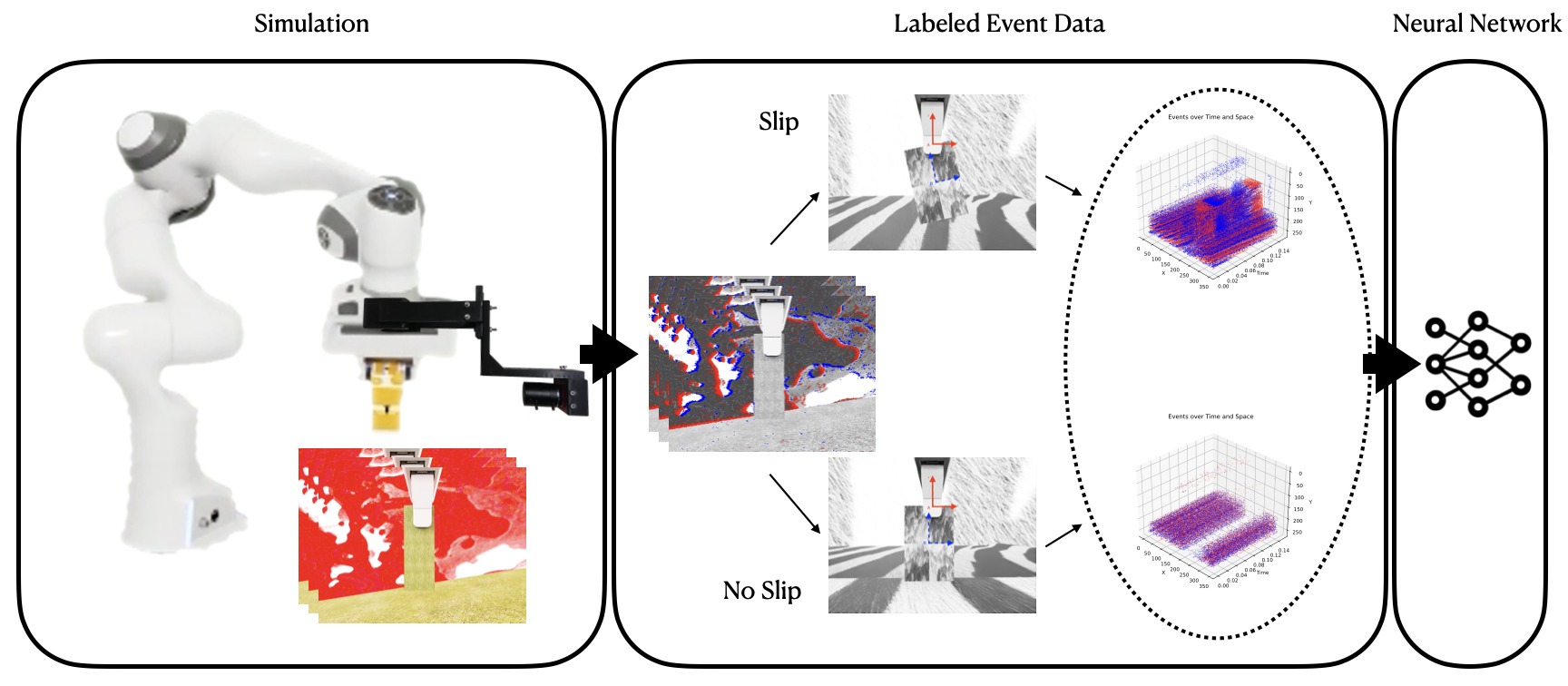}
   \caption{\emph{Overview of the pipeline}: 
   from physics simulation generating synthetic frames of a manipulation task with a Panda robot manipulator (left), through event generation and labeling of slip and non-slip cases (middle), to artificial neural network training (right). 
   The simulation provides ground truth data through the robot's movements, which is used to label the generated event data for supervised learning of slip detection.
   \label{fig:eye_catcher}
   \vspace{-2ex}
   }
\end{figure}

Pick-and-place tasks are a common application for robot manipulators, for example, in industrial manufacturing. Typically, these tasks involve a robot manipulator or arm equipped with a gripper and an object. The task is to grasp the given object with this gripper and move it to a designated position where it is released again. While the task involves several challenges such as determining grasp points, navigating obstacles, and implementing object detection, our work focuses on task observation as a crucial requirement for error handling. The observation can be performed using various sensors such as tactile sensors, cameras, lasers, or event cameras. 
The potential errors range from grasping the wrong object or failing to grasp entirely, to colliding with obstacles, deforming objects, or experiencing slips. A slip occurs when there is unwanted movement of the object while in the gripper's grasp. 
Specifically, a rotational slip involves movement around an axis, while a translational slip involves movement along an axis. 
Rotational slips commonly occur when the grasp point is not precisely aligned in the gravity direction with the object's center of mass. 
For example, when a box with an off-center mass distribution is grasped at its geometric center, it may rotate toward its center of mass if the friction between gripper and object is insufficient. While this rotation alone is problematic, any subsequent movement with the misaligned object can lead to further errors. Tasks like placing the rotated object or lifting it over obstacles may fail due to this misalignment. 
In severe cases, particularly under high accelerations, the grasped object may completely detach or be ejected at high velocities, potentially causing damage.

\begin{figure}[t]
  \centering
   \begin{subfigure}{\linewidth}
        \centering
       \begin{subfigure}{\linewidth}
            \centering
            \includegraphics[width=0.18\linewidth]{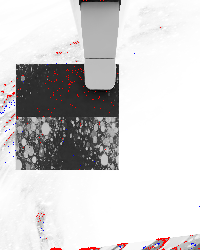}\hspace{1mm}
            \includegraphics[width=0.18\linewidth]{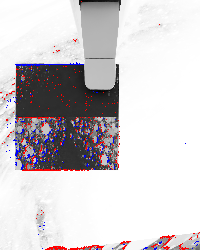}\hspace{1mm}
            \includegraphics[width=0.18\linewidth]{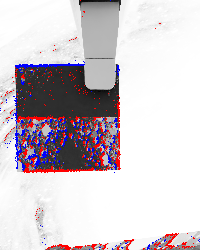}\hspace{1mm}
            \includegraphics[width=0.18\linewidth]{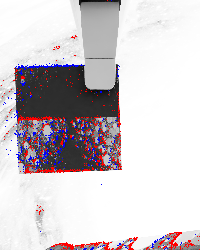}\hspace{1mm}
            \includegraphics[width=0.18\linewidth]{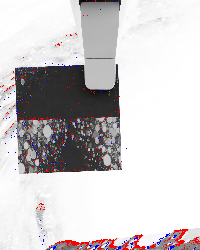}\hspace{1mm}
        \end{subfigure}\hfill
        \begin{subfigure}{1.0\linewidth}
            \centering
            \includegraphics[height=0.38\linewidth]{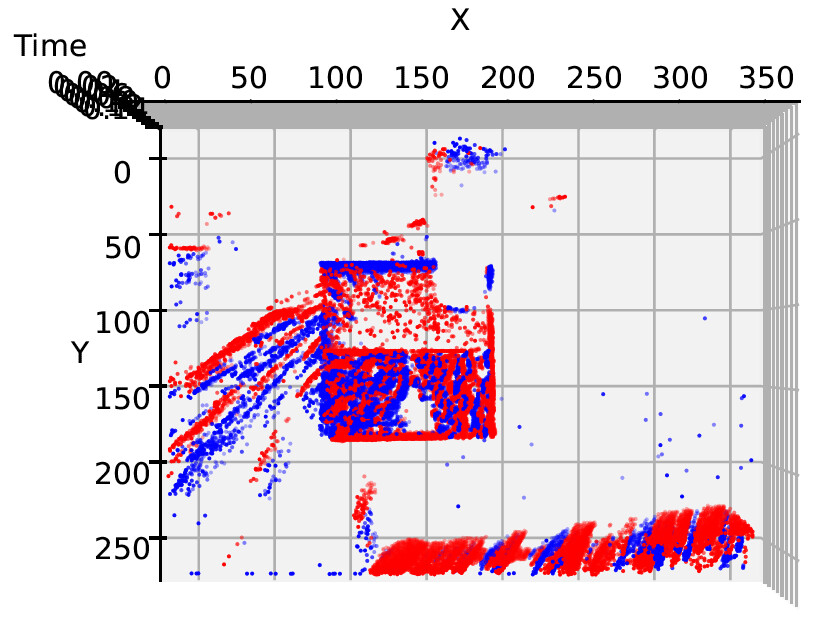}
            \includegraphics[height=0.38\linewidth]{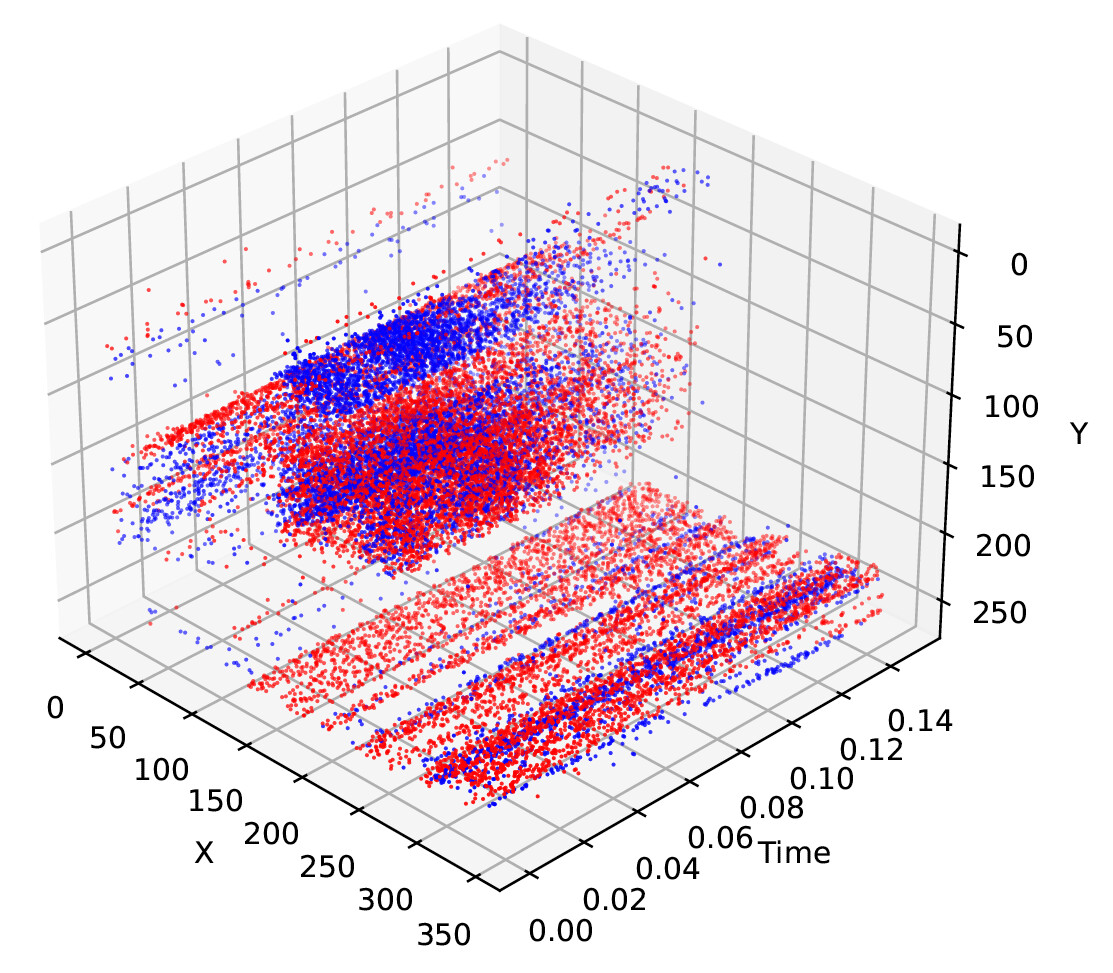}
        \end{subfigure}
        \subcaption{\textbf{Slip}. 
        Top: Frames and events vs.~time. 
        Bottom: 3D plots of events in space-time. 
        Events produced by the object can be seen on the whole interval.}
    \end{subfigure}\\[1ex]
   \begin{subfigure}{\linewidth}
        \centering
        \begin{subfigure}{\linewidth}
            \centering
            \includegraphics[width=0.18\linewidth]{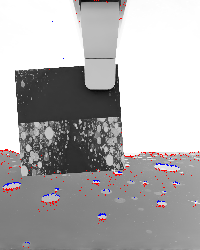}\hspace{1mm}
            \includegraphics[width=0.18\linewidth]{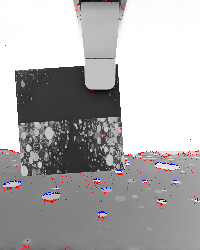}\hspace{1mm}
            \includegraphics[width=0.18\linewidth]{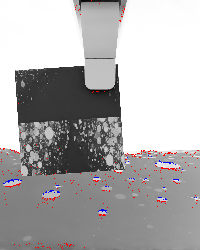}\hspace{1mm}
            \includegraphics[width=0.18\linewidth]{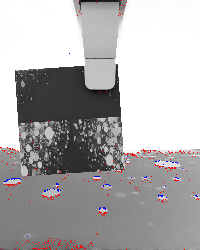}\hspace{1mm}
            \includegraphics[width=0.18\linewidth]{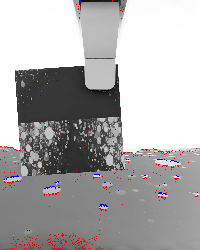}\hspace{1mm}
        \end{subfigure}\hfill
        \begin{subfigure}{0.9\linewidth}
            \centering
            \includegraphics[height=0.38\linewidth]{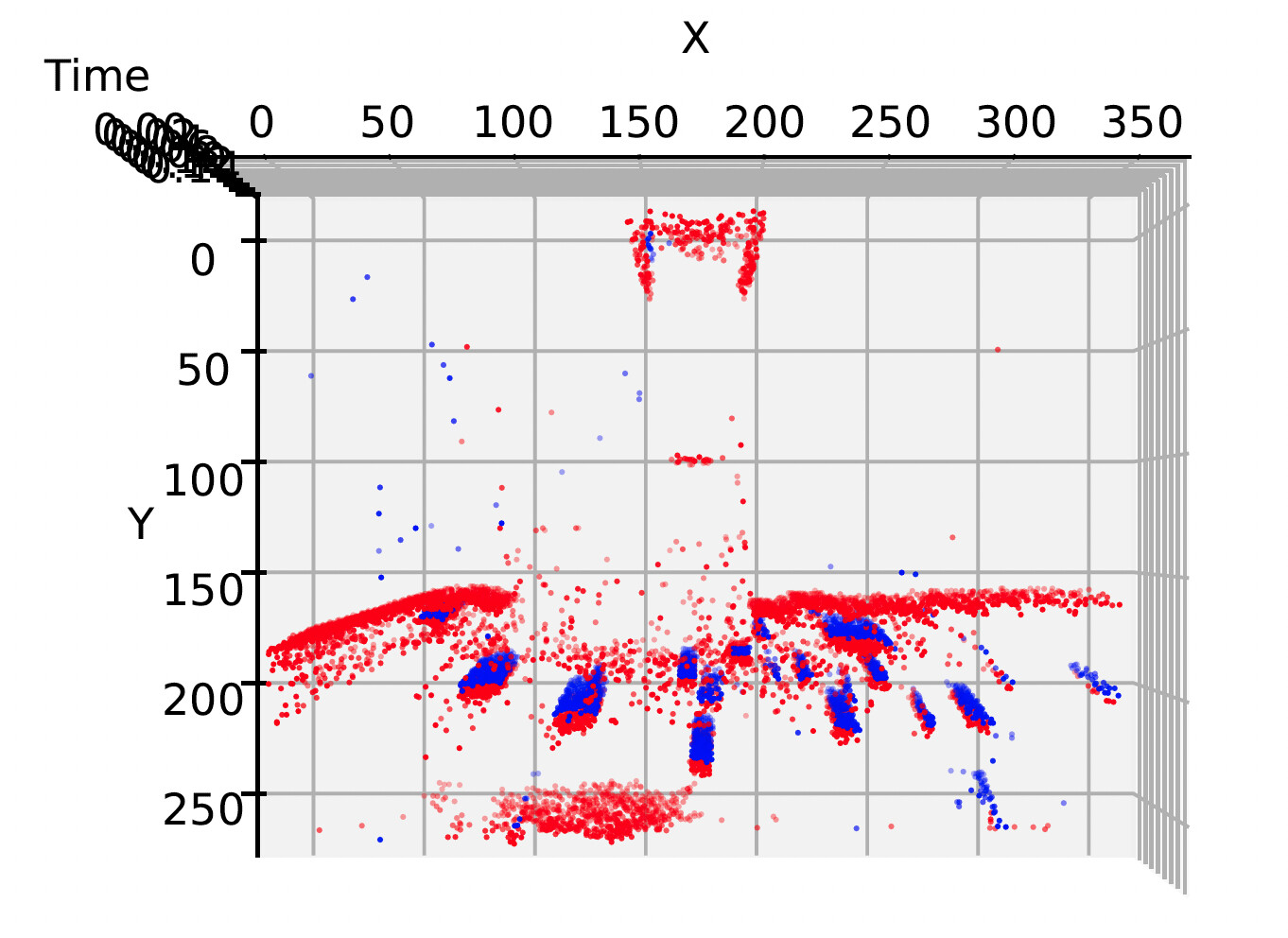}
            \includegraphics[height=0.38\linewidth]{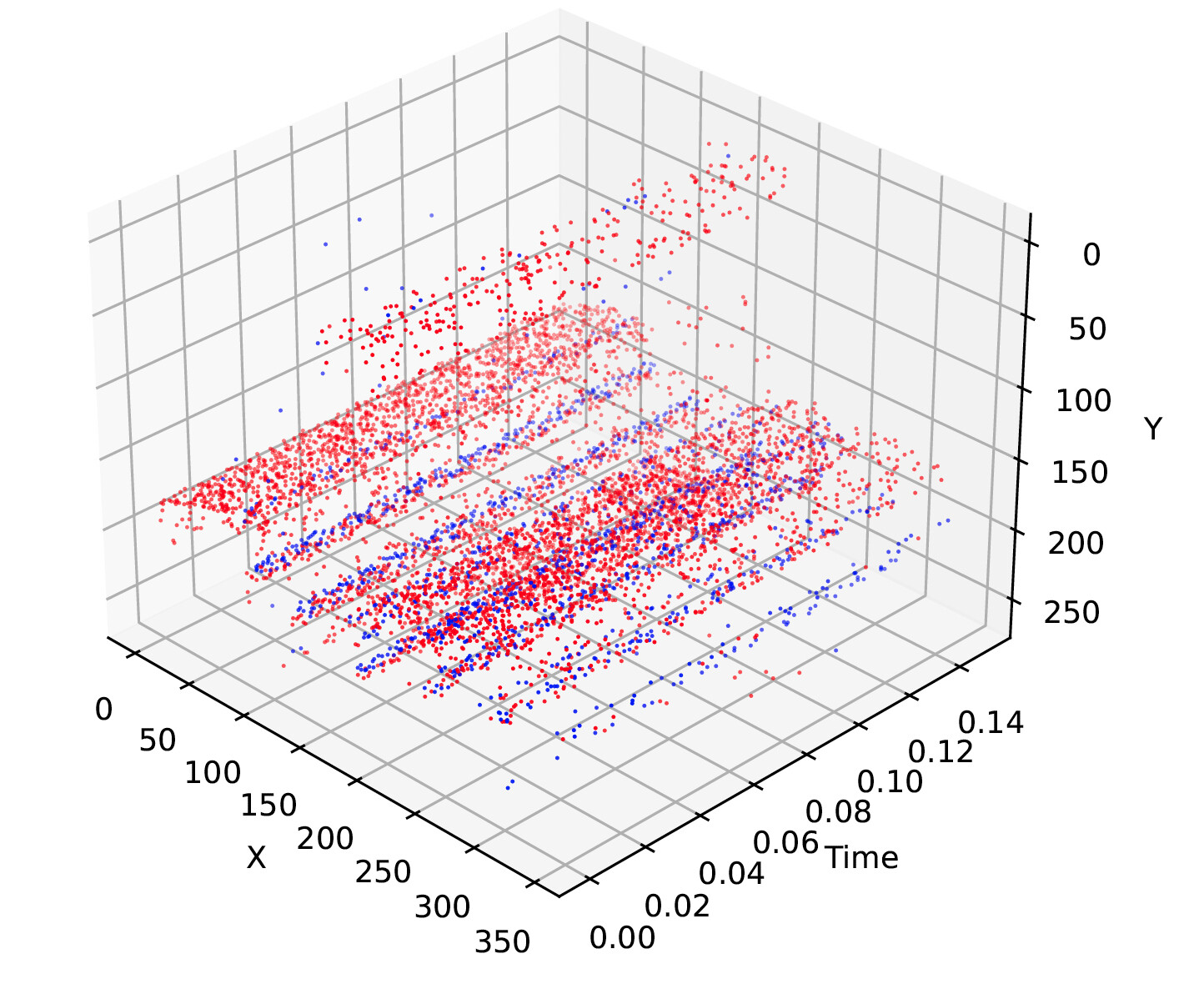}
        \end{subfigure}
        \subcaption{\textbf{No slip}. 
        Top: Frames and events vs.~time. 
        Bottom: 3D plots of events in space-time. 
        No events appear on the object.}
   \end{subfigure}
   \caption{Visualization of two samples of the same simulation scenario, with or without slip.}
   \label{fig:combination_sets}
\end{figure}

This work focuses on detecting %
slips in pick-and-place tasks using event cameras on a robot arm (\cref{fig:eye_catcher}). 
An event camera offers several beneficial features for this task \cite{Lichtsteiner08ssc,Gallego20pami}: high temporal resolution, high dynamic range, and low power consumption. 
However, this paper emphasizes the camera's fundamental operating principle --generating events only when pixel brightness changes-- as the key advantage for slip detection. 
Due to this property, events are only produced by movements relative to the camera, and when the camera is rigidly mounted to the gripper of the robot arm, only movements relative to the gripper generate events. Thus, an object held by the end effector will only produce events if it slips (see \cref{fig:combination_sets}).

In the work of \cite{albert}, a similar approach was explored using an event camera for slip detection. 
While their analytical methods successfully detected slips in specific scenarios, the challenge of parameter fine-tuning for generalization remained unsolved. 
This motivated us to explore data-driven approaches, which would mitigate parameter tuning. 
Upon reviewing related literature, we identified that collecting appropriate data represents one of the primary challenges in addressing this task. 
Having complete control over data quantity and complexity without excessive manual effort provides a significant advantage. 
This enables analytical approaches to start with simple setups and incrementally increase complexity, while also supporting data-driven methods through the generation of large, diverse datasets for both general and specific tasks. 
A simulator provides the means to achieve this control over data generation. It enables full control over scenarios (up to some approximation of real-world conditions), allowing for repeatable experiments with controlled variations and straightforward generation of ground truth data. 

This paper presents a simulation pipeline that generates synthetic event data from photorealistic images, produced by physically accurate simulations of a basic pick-and-place task. 
To the best of our knowledge, it is the first time that such a simulation system is proposed.
The system pipeline simulates a Franka Emika Panda robot manipulator, with a rigidly attached event camera, interacting with a box-like object. It allows modifications of object properties (shape, size, weight), textures, environmental lighting, and gripping positions to control dataset complexity. 
Leveraging the event-camera's principle of detecting only relative motion, our pipeline generates synthetic data specifically suited for slip detection (\cref{fig:eye_catcher}). 
To validate both this approach and the quality of the simulated data, we trained three artificial neural network (ANN) architectures  (see \cref{sec:neural_networks}) on the synthetic data and a real-world dataset, achieving assuring results on the slip detection task.

\section{Related Work}
\label{sec:realted_work}

\subsection{Robot Simulation}
The use of simulators in robotics research has been well established; a notable example is the work of Shariq Iqbal \etal \cite{frame_simulator}, who developed a photorealistic simulator with real-world physics to investigate \emph{directional semantic grasping}. 
This task involves a robot grasping specific objects from predetermined directions within a shelf containing multiple items. 
The researchers successfully demonstrated sim-to-real transfer by training a deep reinforcement learning algorithm for closed-loop control of object grasping on simulated data and deploying it in real-world scenarios.
While these results were promising, the authors acknowledged that their approach remained confined to specific use cases, and broader, more generalized solutions required further research. 
Moreover, their simulator was developed exclusively for conventional (\ie, frame-based) vision sensors and has not been made available to the research community.

Commercial and open-source physics simulators, such as NVIDIA's Isaac Sim \cite{isaac_sim} and PyBullet \cite{pybullet}, provide sophisticated simulation environments. 
These platforms use game engines like \emph{Bullet} and \emph{PhysX} to deliver realistic physics simulations and photorealistic rendering capabilities.

PyBullet and Isaac Sim support robotic simulations, synthetic data generation, and various machine learning applications through their Python interfaces, but they miss event-based data. 
Also, to use these simulators for slip detection, the scenario must first be set up by configuring the environment, defining object properties, integrating a robotic manipulator, and implementing the required sensors.

\subsection{Event-Based Slip Detection}
Recent years have seen numerous publications in the field of event-based cameras, with several focusing on manipulation tasks \cite{Gallego20pami}. 
However, only a limited number of these works address in-hand tracking or slip detection.

Taunyazov \etal \cite{VT-SNN} implemented a multi-modal approach combining two event-based sensors: a NeoTouch tactile sensor and an event camera. 
Their Visual-Tactile Sensing system (VT-SNN) integrates data from both sensors to train a neural network for two distinct tasks: object detection, which distinguishes between visually identical objects of different weights, and rotational slip detection on a single object. 
While their results demonstrate the potential of this approach, the evaluation was limited by a small dataset, lack of object diversity, and brief testing intervals.

Niklas Funk \etal \cite{evetac} introduced an event-driven approach using their optical tactile sensor, \emph{Evetac}. The sensor consists of a soft, non-transparent silicone gel with embedded markers attached to the gripper, where deformations from object contact and movement are captured by an event camera positioned behind the gel. 
While similar optical tactile sensors exist, they traditionally use frame-based cameras. 
The system achieved impressive results with slip detection at 1kHz, albeit in a constrained scenario with limited object variety and no manipulator movement. 
The sensor's independence from environmental motion and lighting conditions significantly reduced the dataset complexity requirements. 
The work stands out for its reproducibility through commercial components, 3D-printable parts, and open-source algorithms, though it requires time-consuming manual grasp execution during data collection and gel preparation for ground-truth sessions.

Rajkumar Muthusamy \etal \cite{baxter} developed an event-based finger vision system for slip detection and suppression during manipulation tasks. Their system employs a parallel gripper with translucent finger plates and an event camera. 
While achieving a high slip detection rate of 2kHz, the system faces several constraints: a highly restricted testing environment, limited object movements, dependence on sufficient object surface texture, and the requirement for manual data collection.

In \cite{albert}, various analytical approaches to rotational slip detection are investigated with a setup similar to the one we consider, by using the event rate, event-based optical flow \cite{Zhu18rss,Shiba22eccv}, etc. %
The methods exploit the principle that objects generate events only during relative motion between the camera and gripper. 
Compared to previous approaches, this work examines longer manipulation periods with more dynamic movements. 
However, the camera viewpoint presents a limitation, as rotational slip movements toward or away from the camera are difficult to detect. Observing from a lateral perspective could improve the detection of such movements. 
The authors concluded that more diverse scenarios were needed for robust and generalizable results.

The reviewed works demonstrate promising approaches to event-based slip detection but share common limitations in their evaluation. 
A recurring challenge is the time-consuming nature of data collection, with most approaches requiring manual execution of grasps and complex setup procedures. Additionally, the existing studies are typically restricted to controlled environments with limited object variety and constrained movements. While \cite{albert} examines longer manipulation periods, all approaches would benefit from more diverse scenarios to achieve robust and generalizable results. These limitations highlight the need for a systematic approach to generate comprehensive datasets for event-based slip detection. A simulation environment could address these challenges through automated data generation in various scenarios, objects, and grasp conditions.

\section{Simulation and Slip Detection Pipeline}
\label{sec:simulation_and_slip_detection_pipeline}

\begin{figure*}[htbp]
    \centering
    \includegraphics[width=0.9\linewidth]{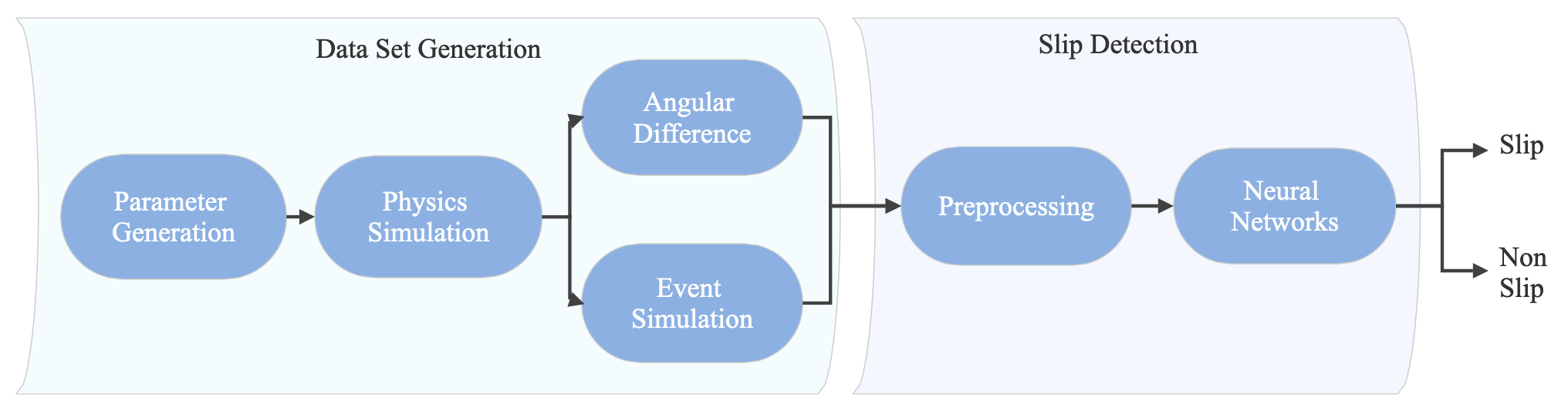}
    \caption{Data simulation and slip detection pipeline.
    \label{fig:flow_chart}
    \vspace{-2ex}
    }
\end{figure*}

We present a complete pipeline that covers both data generation and data-driven slip detection (through ANN training), as illustrated in \cref{fig:flow_chart}. 
The pipeline consists of two independent modules: a data generation system and an ANN training framework. 
These modules can be utilized either as an integrated pipeline or as standalone components.

\subsection{Dataset Generation}
The complete pipeline for the generation of event-based simulated datasets consists of four steps, which we describe in the upcoming sections. 
The first step is the parameter generation for each dataset. 
The second step is the physics simulation, which generates frames and position values of the object and the gripper. 
The third step is the simulation of event camera data. 
The final step is the calculation of the angular difference for every generated sample, which serves as ground-truth (GT) data.

\subsubsection{Parameter Generation}
\label{sec:parameter_generation}
The input for this step determines the size of the dataset, whether the dataset is split into training and test sets, and how the parameter values of the simulation scene are selected. 
A dataset consists of individual samples, where each sample contains all the data generated from a simulation. 
The size of the datasets has no constraints but the available computation time is usually the limiting factor. 
A split into training and test sets guarantees that no parameter value of the test set is previously used in the trainings set. 
The parameters for a simulation are either generated by an exhaustive combination of a list of selected parameters or by random selection from a value range (for a detailed description of the parameters see \cref{sec:physics_simulation}).

\subsubsection{Physics Simulation}
\label{sec:physics_simulation}
This step executes the physics simulation to generate gripper and object orientations for ground truth, along with synthetic images (\ie, frames) for event data generation. 
The simulation utilizes NVIDIA's Isaac Sim simulator \cite{isaac_sim}, a powerful robotics platform that provides realistic physics and produces photorealistic simulations. 
While Isaac Sim offers a graphical editor for manual scenario building and testing, we implement programmatic control through Python scripts. 
Our approach enables access to all configurable parameters as both inputs and outputs, making it highly versatile for data generation.

\begin{figure}[t]
  \centering
   \begin{subfigure}{.8\linewidth}
        \centering
        \includegraphics[width=\linewidth]{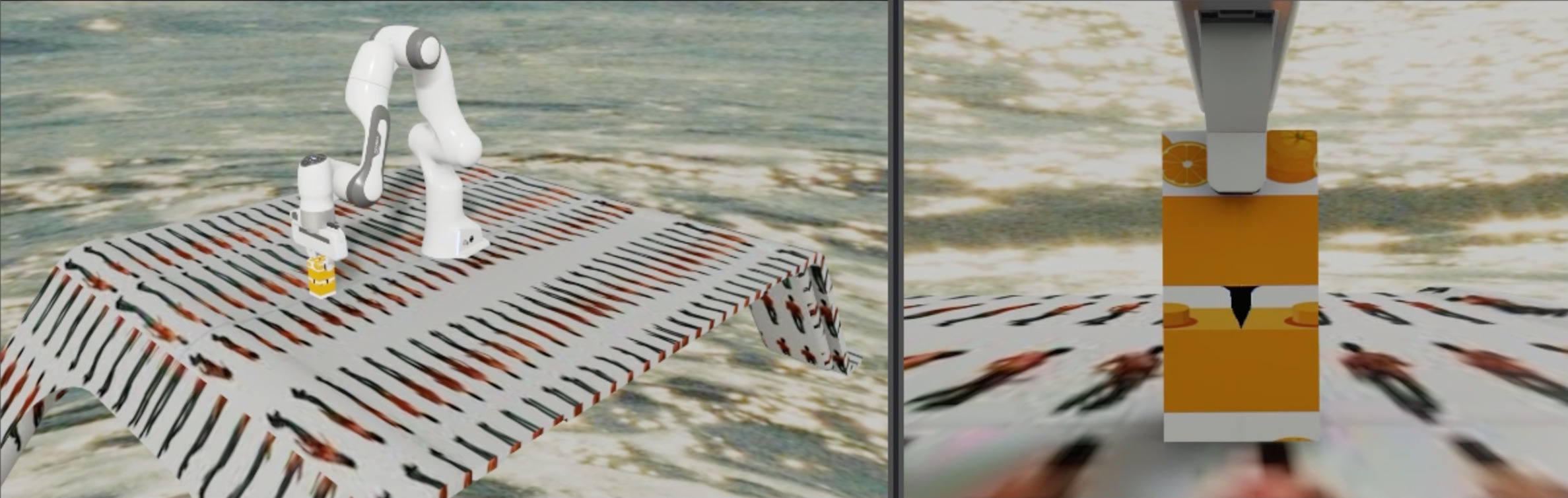}
        \includegraphics[width=\linewidth]{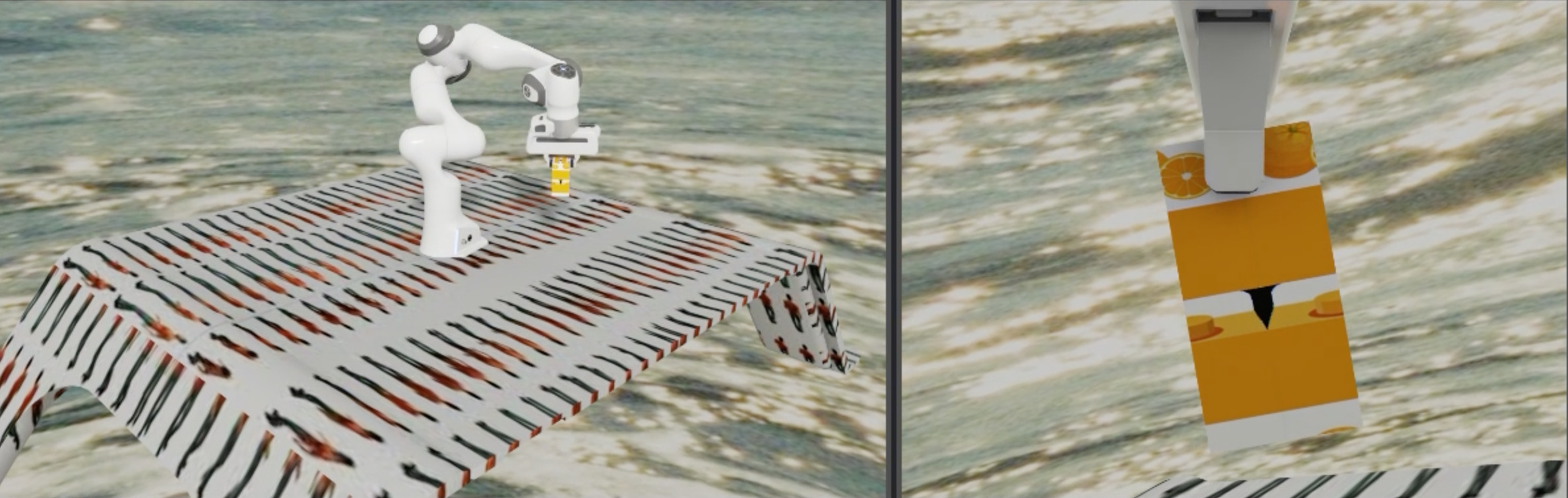}
        \includegraphics[width=\linewidth]{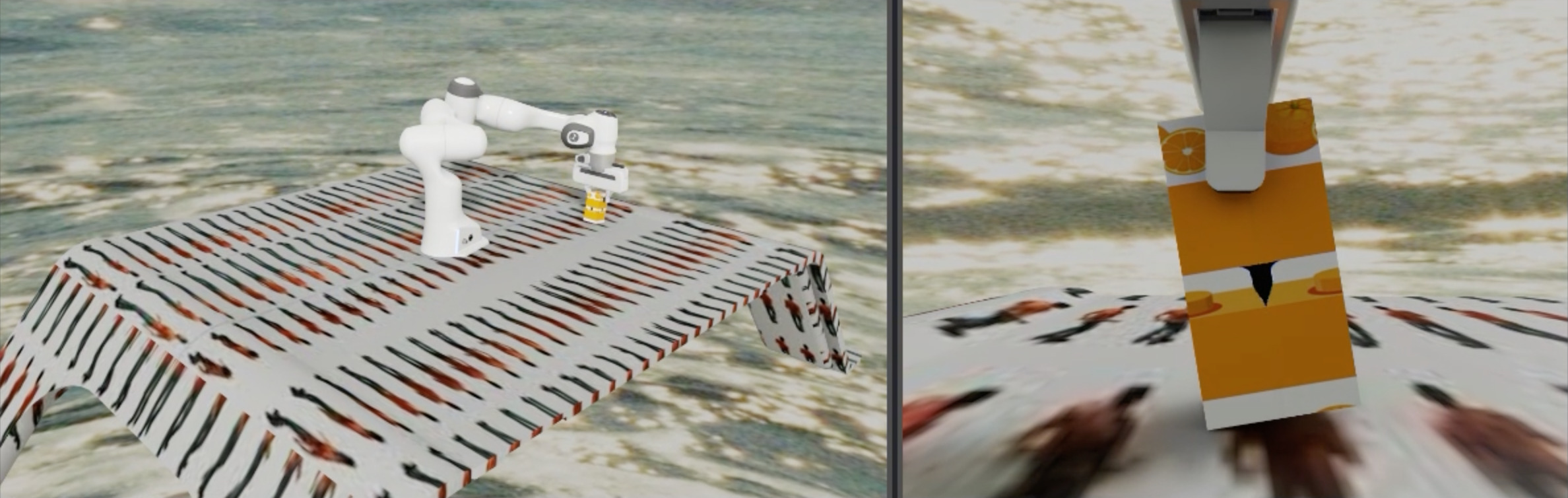}
    \end{subfigure}
    \caption{Simulator Scene. 
    The left side shows the overall scene, and the right side shows the view from the camera attached to the robot arm.
    Execution of a whole pick-and-place task is shown sequentially from top to bottom.
    \label{fig:isaac_scene}
    \vspace{-2ex}
    }    
\end{figure}

The basic scene consists of a Franka Emika Panda robot manipulator mounted on a flat surface, a camera focused on the gripper, and a simple cuboid for pick and place operations. 
A light source is positioned above the scene, and the entire setup is enclosed within a sphere to control background texture and lighting (see \cref{fig:isaac_scene}).
The simulation setup offers various adjustable parameters whose values are generated in \cref{sec:parameter_generation}. 
These include the textures of the cuboid, ground surface, and the inside of the surrounding sphere, as well as lighting conditions through the sphere's emission and the internal light source's brightness and position. The gripping points on the cuboid, the cuboid's width, height, and mass are adjustable.

 be influenced through several parameters like the object's mass, width, and gripping position. 
For example, gripping positions far from the center increase slip probability. 
While there is no direct parameter to toggle slip instances, they can be effectively controlled via parameter combinations. 
Maximum mass settings guarantee slip occurrence, while minimum mass prevents it. 
Intermediate values produce varying results depending on other parameters, \eg, a fixed-mass cuboid might slip when gripped at its edge but remain stable when gripped centrally. 
This approach allows most parameters to generate both slip and no-slip cases while enabling various slip types at different speeds and positions along the trajectory.

\subsubsection{Event Generation}
Event generation is performed using the v2e simulator \cite{Hu21cvprw}, which converts frames into synthetic event-camera data after interpolating between frames using a pre-trained ANN. Since the maximum rendering frame rate of the Isaac Sim simulation was limited to 60Hz, an upsampling mechanism was needed to achieve realistic event generation. The simulated frames, being free of motion blur, are particularly well fit for this upsampling and event generation process.
Through comprehensive noise modeling, the simulator generates events that closely approximate real event camera data. The frames from the Isaac Sim physics simulation serve as input to v2e, along with several configurable parameters.
The key parameters in this work are the event threshold (contrast sensitivity of the event generation model \cite{Gallego20pami,Gallego15arxiv}), input frame rate (60Hz), and output resolution of 346$\times$ 260 pixels, which matches the specifications of the DAVIS346 camera \cite{davis346}. While additional v2e parameters can be adjusted as needed, these were identified as the most critical for our application.
The simulation produces two output files: an event data file where each row contains an event's timestamp, $x$-coordinate, $y$-coordinate, and polarity, and a parameter file documenting the v2e configuration used in the simulation.

\subsubsection{Angular Difference}
\label{sec:ang_diff}
\begin{figure}[t]
\def\figHeight{0.42\linewidth}
  \centering
   \begin{subfigure}{\linewidth}
        \centering
        \begin{subfigure}{.9\linewidth}
            \centering
            \includegraphics[height=\figHeight]{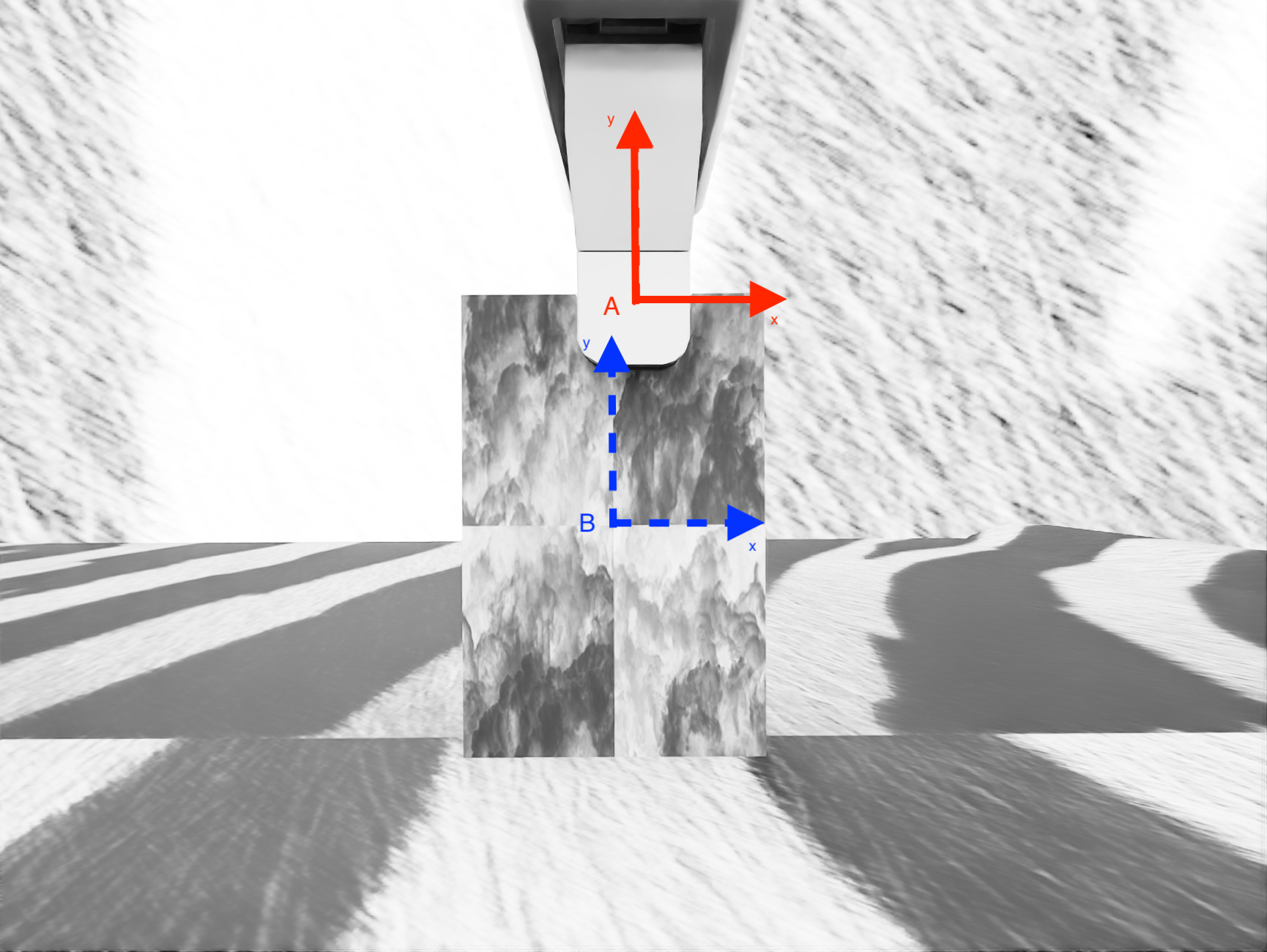}
            \includegraphics[height=\figHeight]{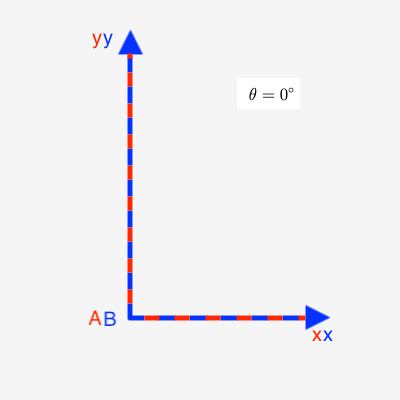}
            \subcaption{Start of the sequence, prior to the slip. 
            The orientations of the two coordinate systems agree.}
        \end{subfigure}\\[1ex]
        \begin{subfigure}{.9\linewidth}
            \centering
            \includegraphics[height=\figHeight]{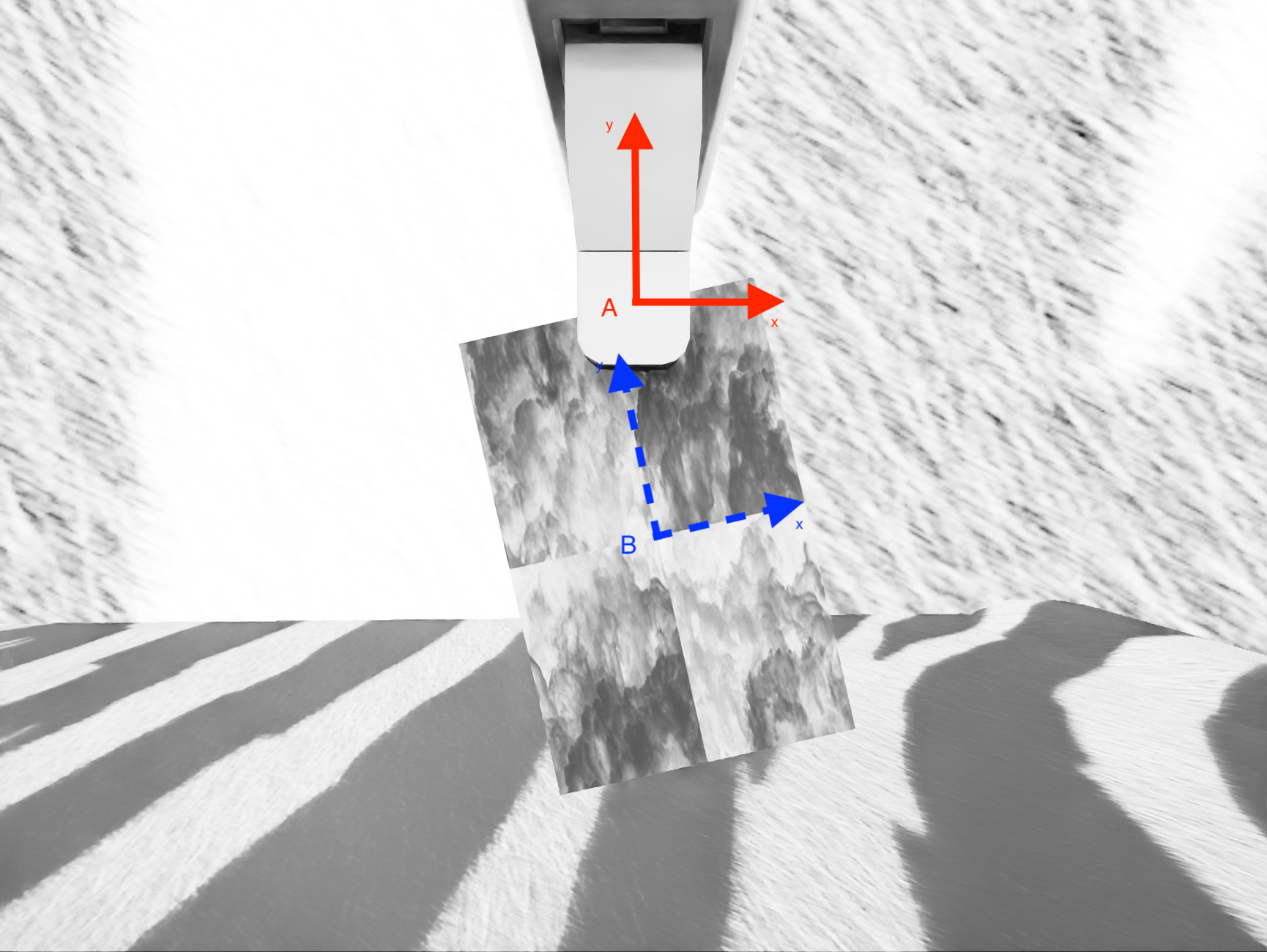}
            \includegraphics[height=\figHeight]{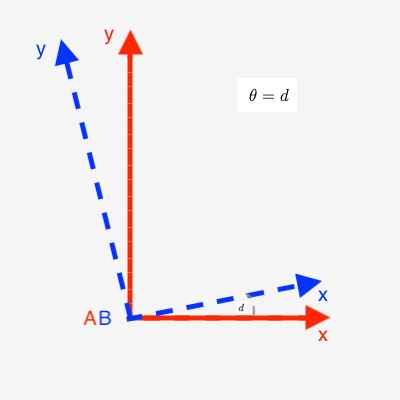}
            \subcaption{Half a second into the sequence, a rotational slip has occurred. 
            The orientations of the two coordinate systems are no longer aligned, as shown by their angular difference.}
        \end{subfigure}
    \end{subfigure}
    \caption{Two images of a slip sequence, 
    with their coordinate systems marked in red and blue, illustrate a rotational slip and the resulting angular difference.
    \label{fig:ang_diff}
    \vspace{-4ex}
    }
\end{figure}

To establish ground truth, we calculate the angular distance between the object and gripper (see \cref{fig:ang_diff}). 
During each simulation step, the positions and orientations of the gripper are computed and recorded.
The initial positions of both the gripper and cube are stored as rotation matrices $B_\text{gripper}$ and $B_\text{cube}$. 
These matrices serve as reference points for calculating subsequent angular differences, enabling measurement of relative changes from the initial configuration.
The angular difference $\theta$ (in radians) is calculated at each iteration using the following formulas:
\begin{equation}
\label{eq1}
\begin{split}
& \Delta M_\text{gripper} = M_\text{gripper}\cdot B_\text{gripper}^\top \\
& \Delta M_\text{cube} = M_\text{cube}\cdot B_\text{cube}^\top \\
& \theta = \left| \arccos \Bigl(\frac{\textrm{tr}(\Delta M_\text{gripper}^\top\cdot \Delta M_\text{cube}) - 1}{2}\Bigr)\right|
\end{split}
\end{equation}
where $M_\text{gripper}$ and $M_\text{cube}$ represent the rotation matrices that describe the current orientation of the gripper and cube, respectively.
The classification of samples as slip or non-slip requires careful consideration of angular differences $\theta$. 
In this work, two thresholds were defined: $\theta_\text{slip}$ for the minimum angular difference indicating slip, and $\theta_\text{non-slip}$ for the maximum allowed angular difference in non-slip cases. 
Defining these thresholds is challenging as angular changes can occur even without visible slippage, due to factors such as camera resolution limitations, extremely slow movements, or camera perspective. 
These subtle changes might not be perceptible to the human eye, yet they still register as measurable angular differences.

\subsection{Slip Detection}
\subsubsection{Preprocessing}
\label{sec:preprocessing}
The preprocessing serves four key objectives: 
($i$) breaking down each sample's event stream into fixed-length subsamples, 
($ii$) generating ground truth labels by categorizing subsamples as slip or non-slip, 
($iii$) optionally visualizing the data to ensure human interpretability, 
and ($iv$) formatting the data to fit the ANN slip classifier.
The subsample length is fixed at 0.16 seconds in this work, balancing rapid slip detection (comparable to VT-SNN's 0.15 seconds) with practical visualization considerations. 
The physics simulator generates frames at 60 Hz (every 0.016 seconds in simulation time), resulting in exactly 10 frames per subsample, streamlining the visualization process.

The ground truth labeling algorithm requires two angular threshold parameters: a slip threshold $\theta_\text{slip}$ that defines the minimum angular difference to classify a subsample as slip, and a non-slip threshold $\theta_\text{non-slip}$ that defines the maximum angular difference to classify a subsample as non-slip (see \cref{sec:ang_diff}). 
When these thresholds are set to the same value, the dataset is cleanly divided into slip and non-slip categories. 
Using different values creates an exclusion zone: subsamples with angular differences falling between the thresholds are omitted from the dataset. This exclusion can lead to clearer decision boundaries during ANN slip-detection training.

For each subsample, videos and individual frames, both with and without events, are created for visualization purposes. The visualization can be deactivated through input settings to save computational time and disk space.
To match the network architectures of \cite{VT-SNN}, spatial data is cropped from 346$\times$260 px to 200$\times$250 px, which also helps focus on the region of interest by reducing background events. 
For each subsample, a .mat file is generated, ready for use with ANNs for slip detection. 
These files contain the events and are named with the label - rotation.mat or stable.mat. 
To ensure balanced data, slip and non-slip subsamples are counted separately, and random samples from the larger subset are removed until both categories contain equal number of subsamples.

\subsubsection{Artificial Neural Networks}
\label{sec:neural_networks}
The implementation follows three artificial neural network (ANN) architectures, as configured in \cite{VT-SNN}. 
Each dataset is divided into training and validation sets using a 80:20 split ratio. 
The subsamples, each 0.16 seconds in duration, are processed differently according to the network architecture: 150 time bins for the Spiking Neural Network (SNN) and Multi-Layer Perceptron (MLP) architectures, and 350 time bins for the 3D-CNN (Convolutional Neural Network) architecture. 
The ANNs perform binary classification between slip and non-slip subsamples. 
Two optimizers are employed to train the ANNs, as detailed in the experiments.

\section{Slip Detection Datasets}
\label{sec:slip_detection_data_set}

\begin{figure}[t]
\def\figWidth{0.19\linewidth}
    \centering
   \begin{subfigure}{1.0\linewidth}
        \centering
        \includegraphics[width=\figWidth]{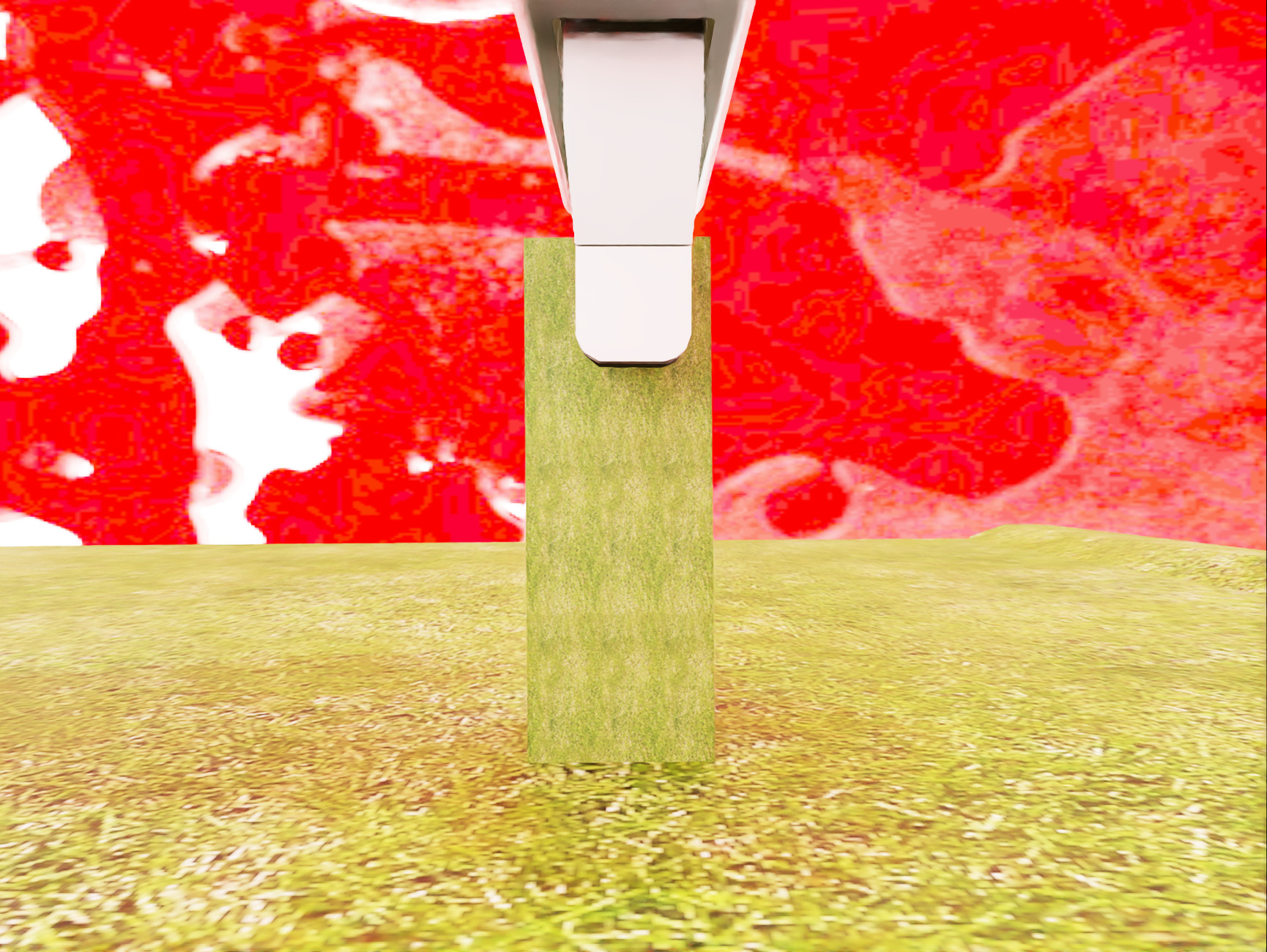}
        \includegraphics[width=\figWidth]{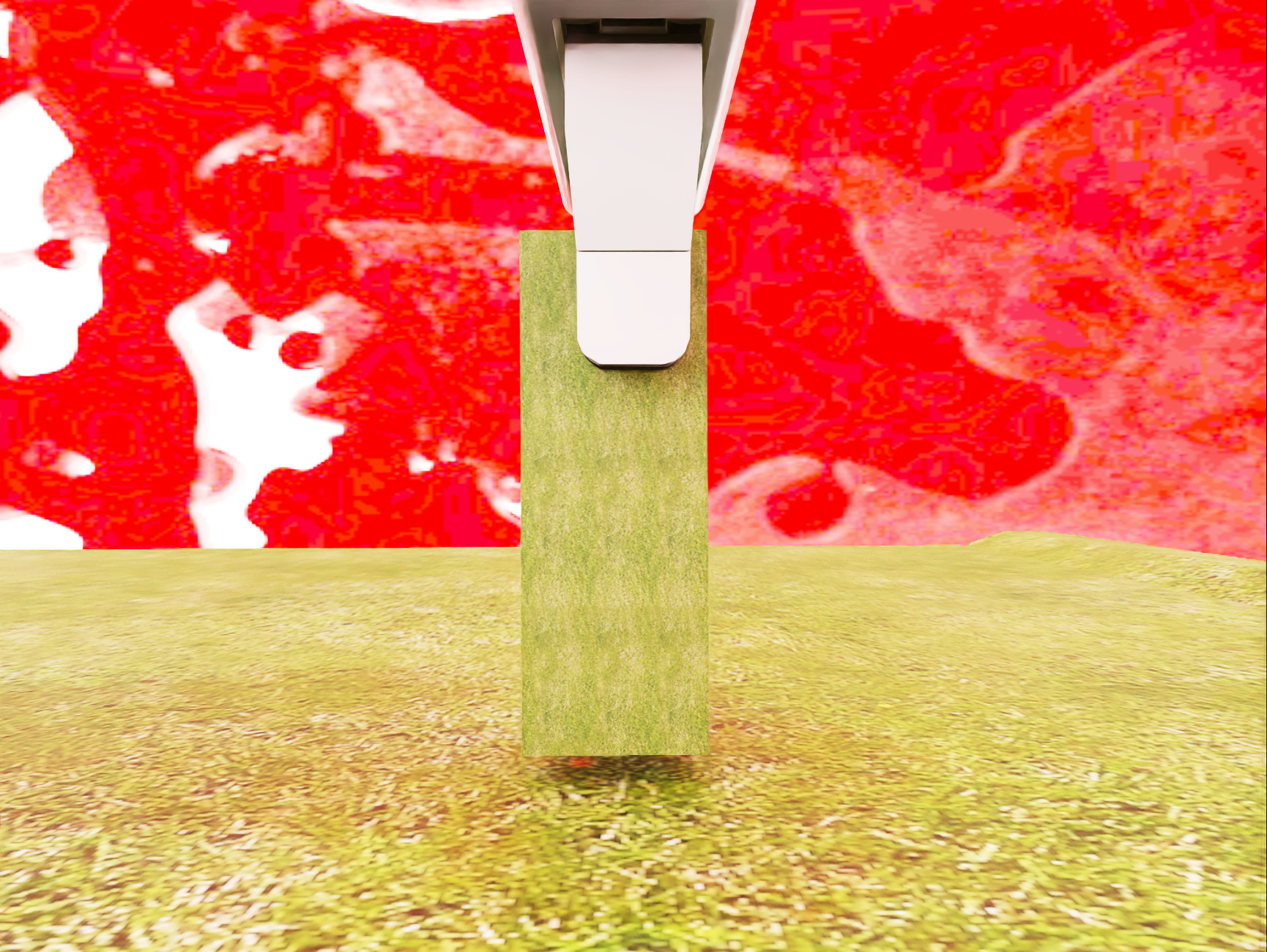}
        \includegraphics[width=\figWidth]{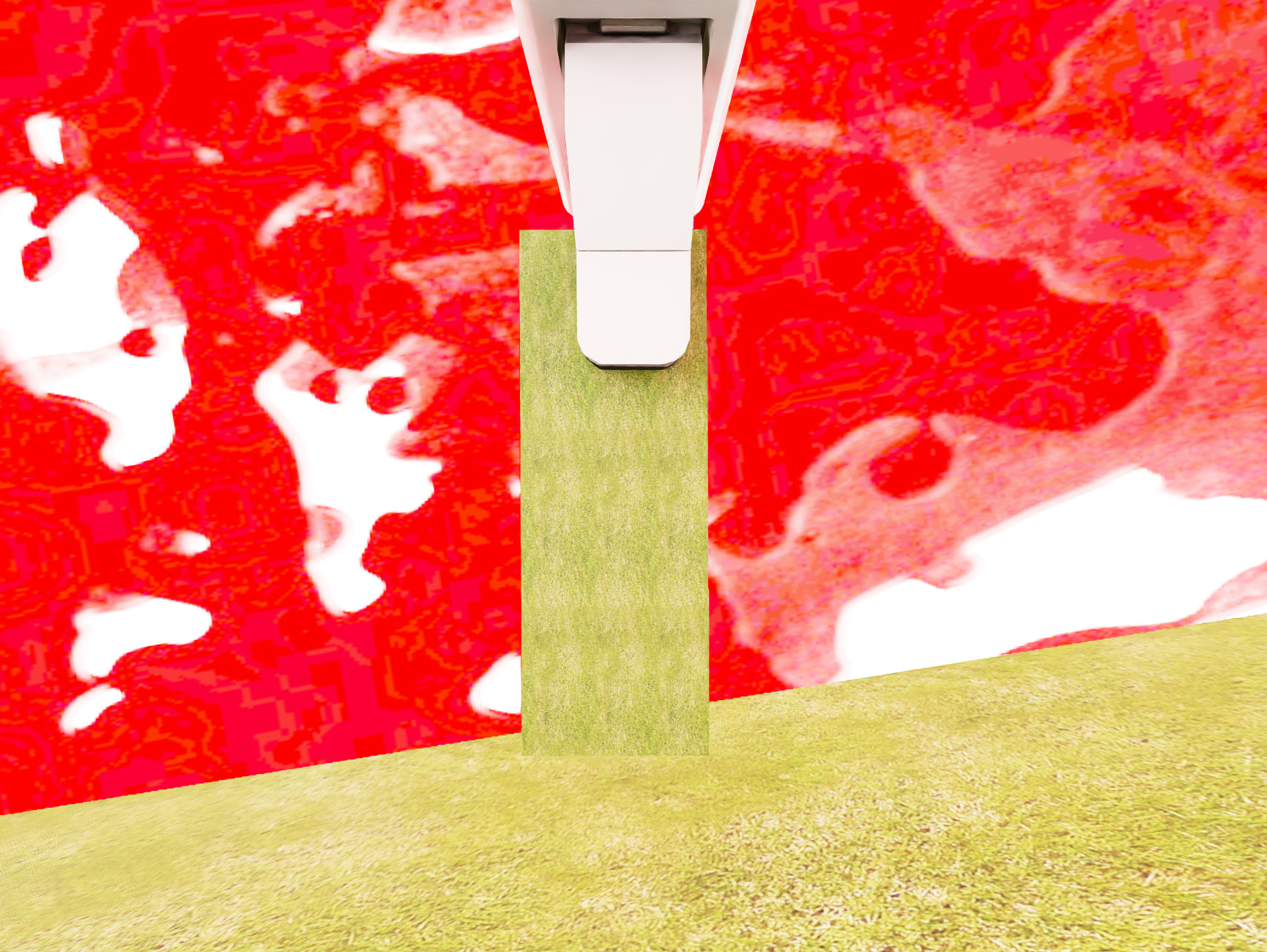}
        \includegraphics[width=\figWidth]{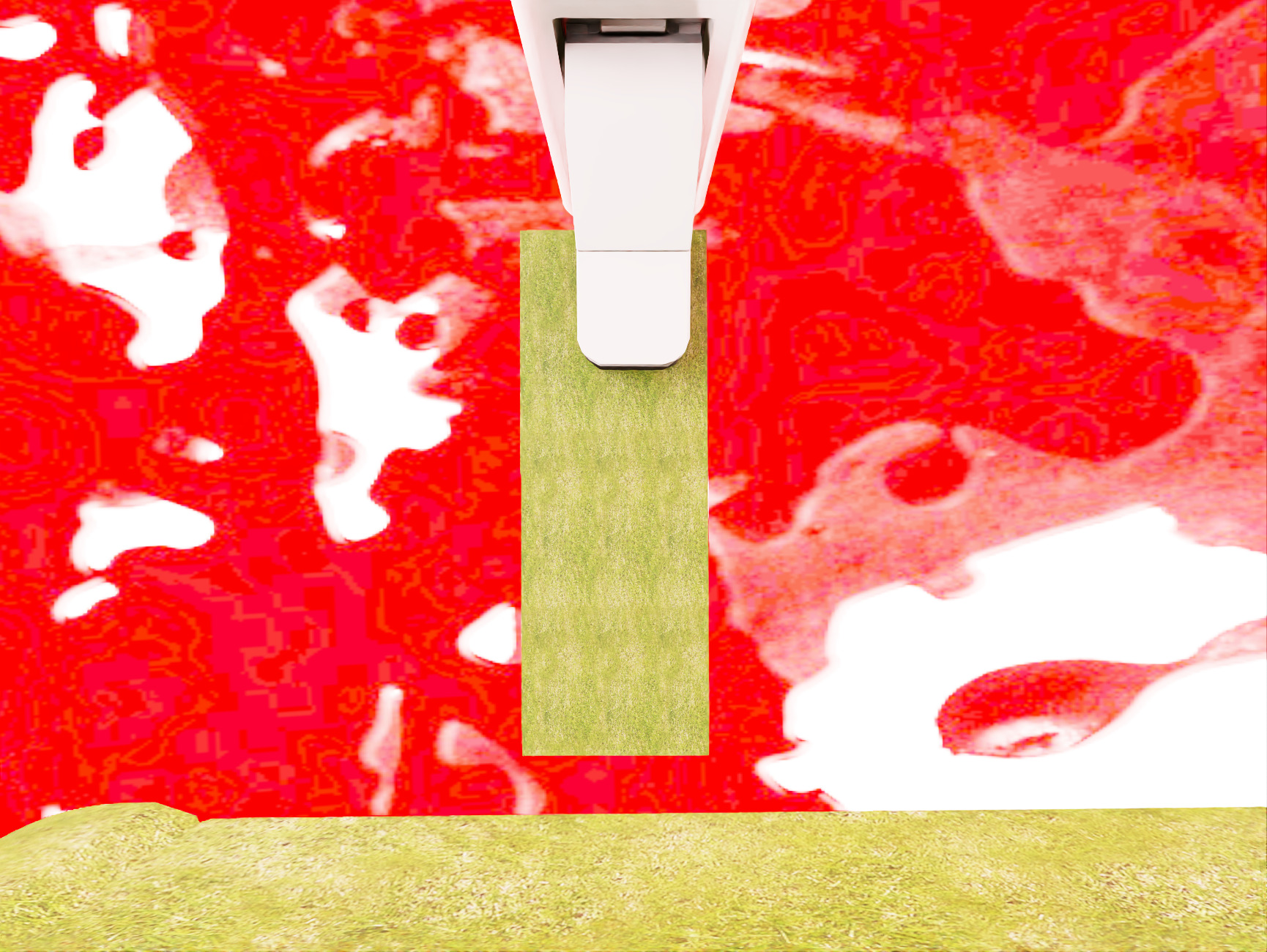}
        \includegraphics[width=\figWidth]{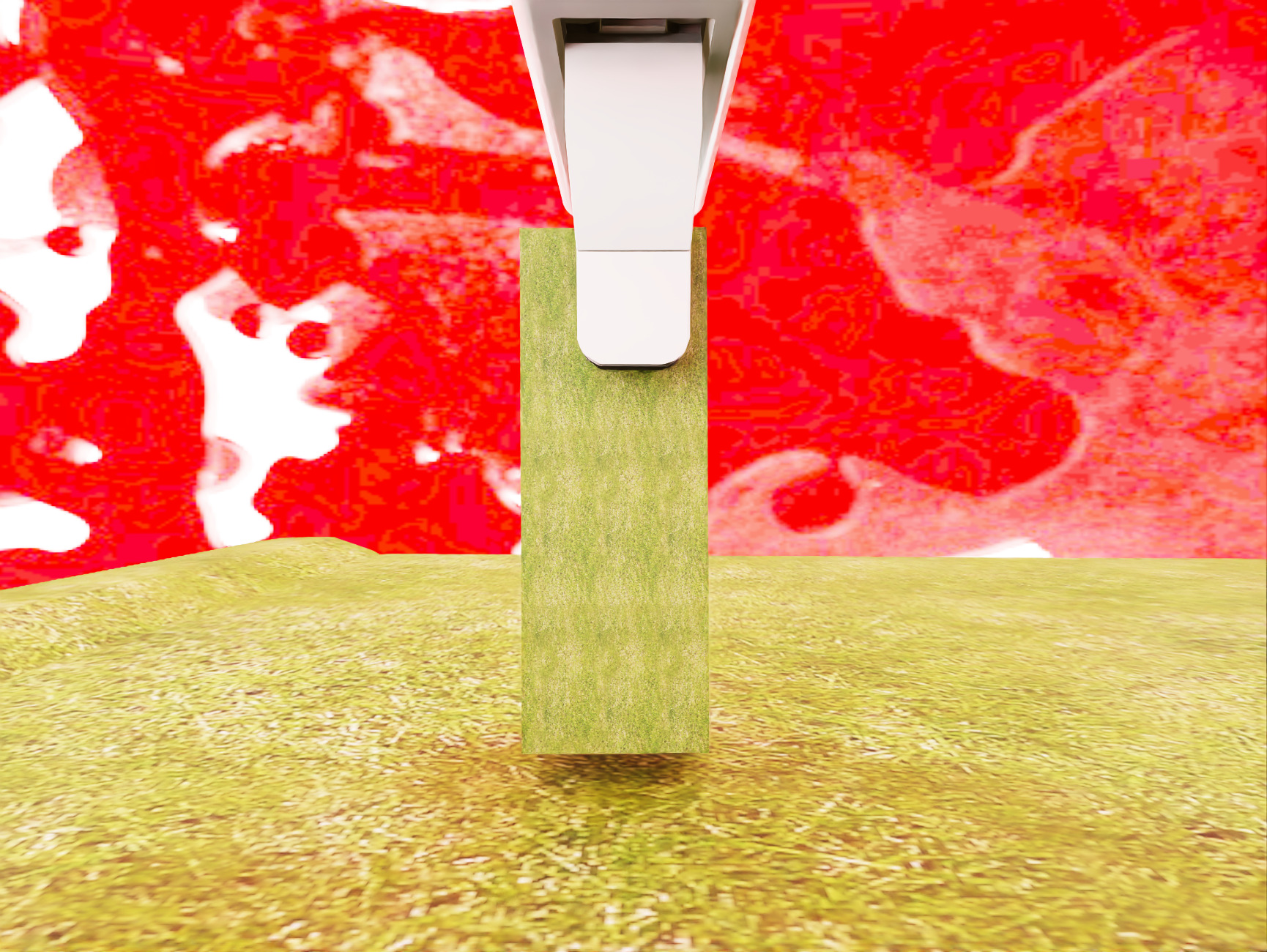}
        \includegraphics[width=\figWidth]{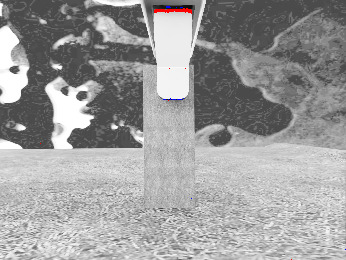}
        \includegraphics[width=\figWidth]{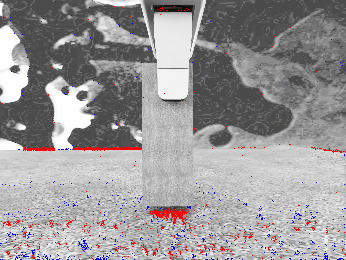}
        \includegraphics[width=\figWidth]{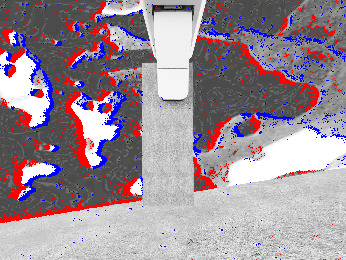}
        \includegraphics[width=\figWidth]{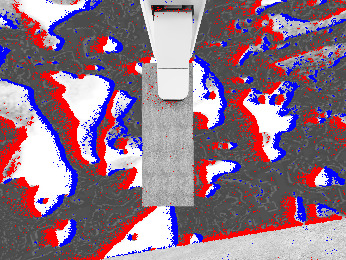}
        \includegraphics[width=\figWidth]{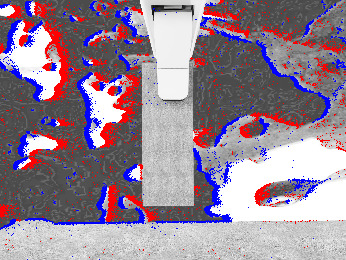}
        \subcaption{Non-slip sample.}
    \end{subfigure}\\[1ex]
    \begin{subfigure}{1.0\linewidth}
        \centering
        \includegraphics[width=\figWidth]{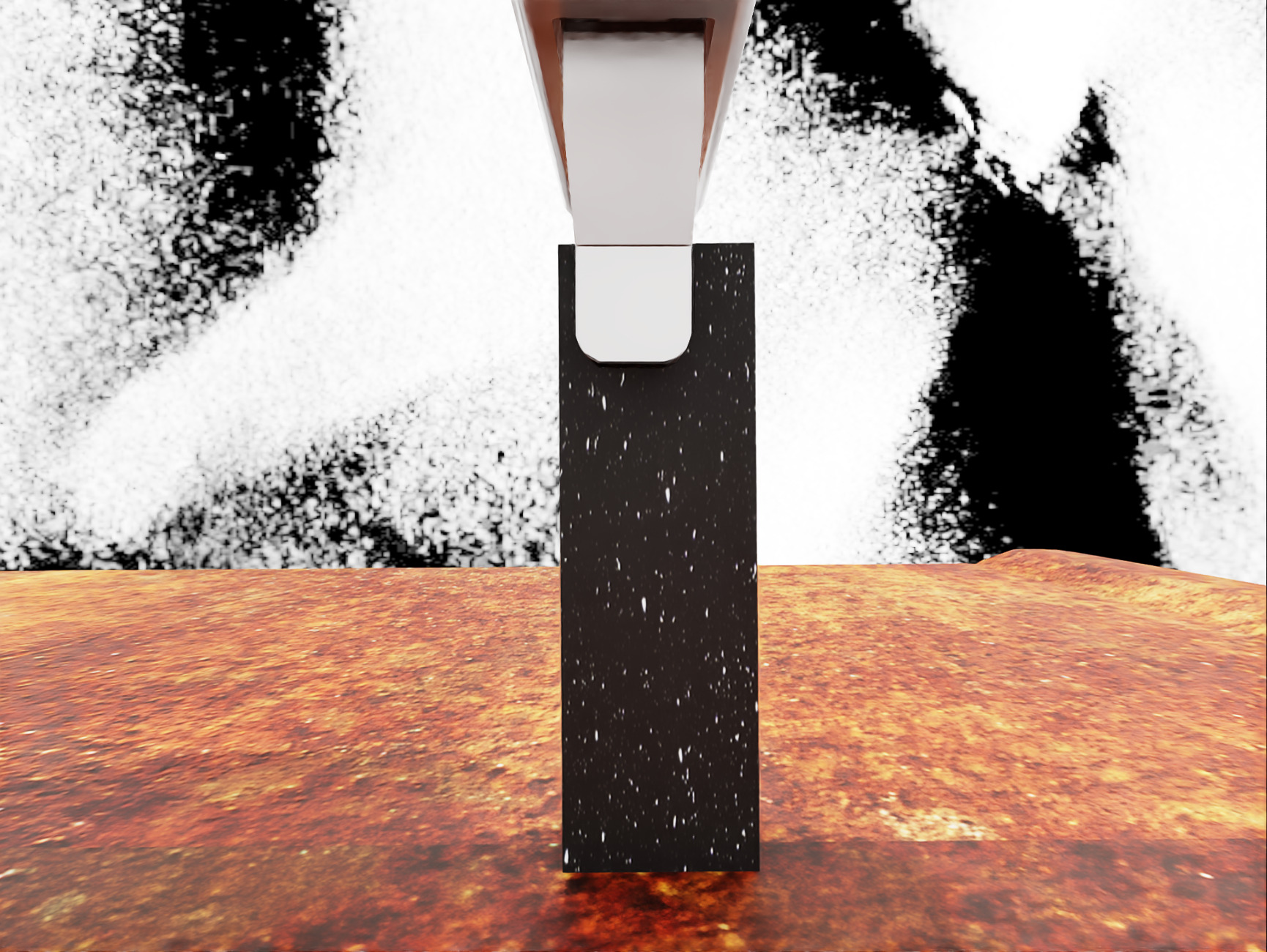}
        \includegraphics[width=\figWidth]{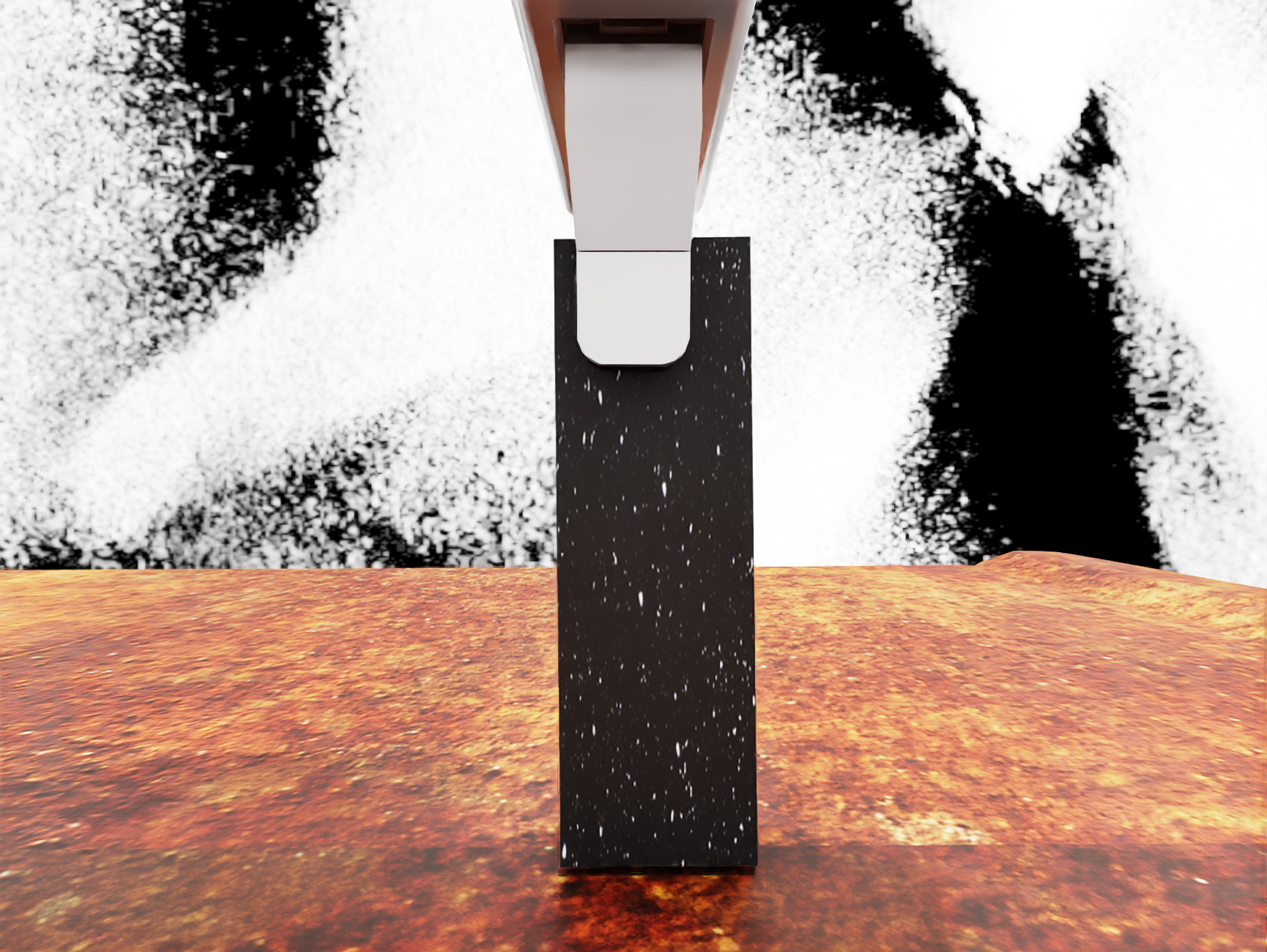}
        \includegraphics[width=\figWidth]{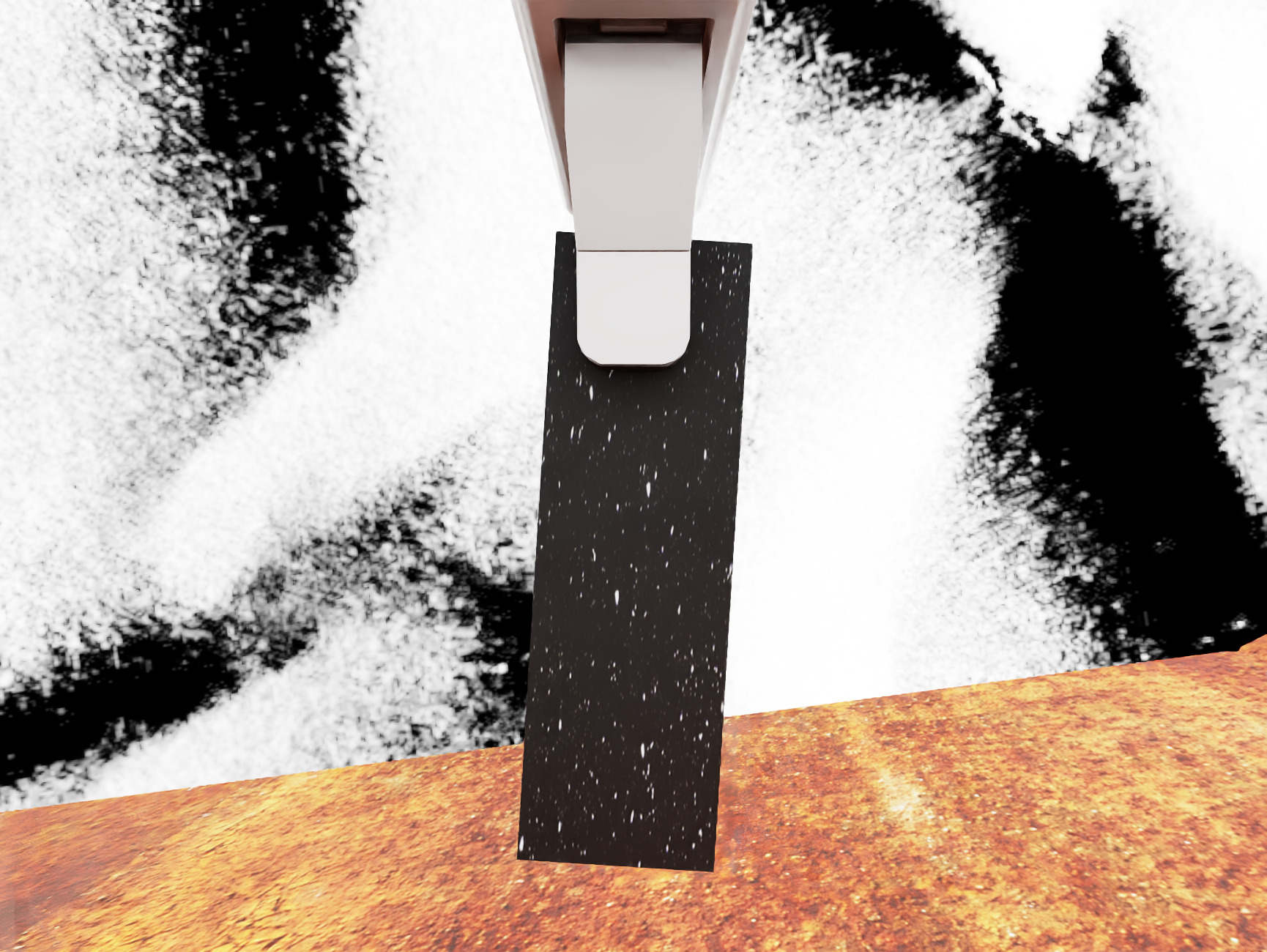}
        \includegraphics[width=\figWidth]{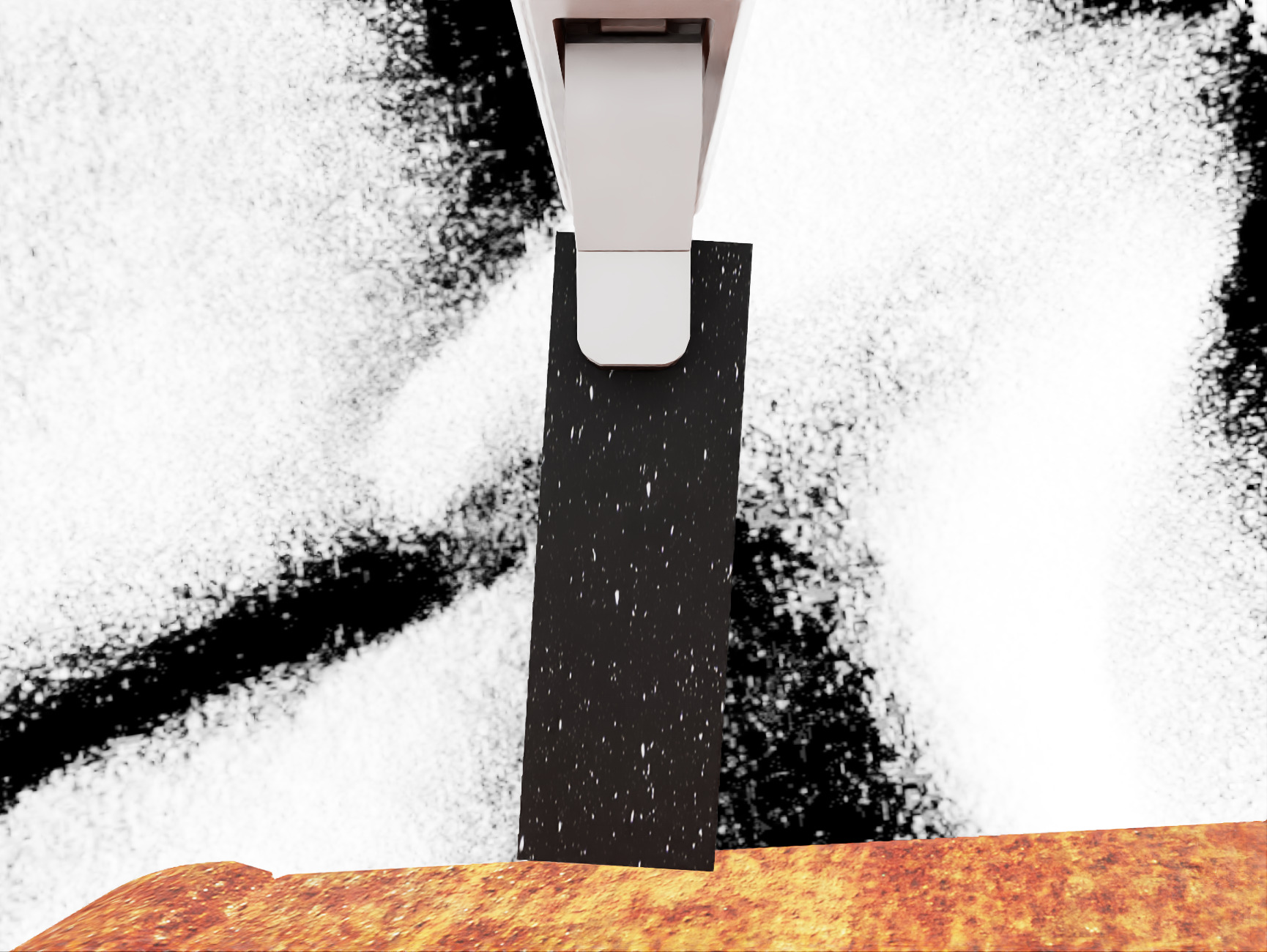}
        \includegraphics[width=\figWidth]{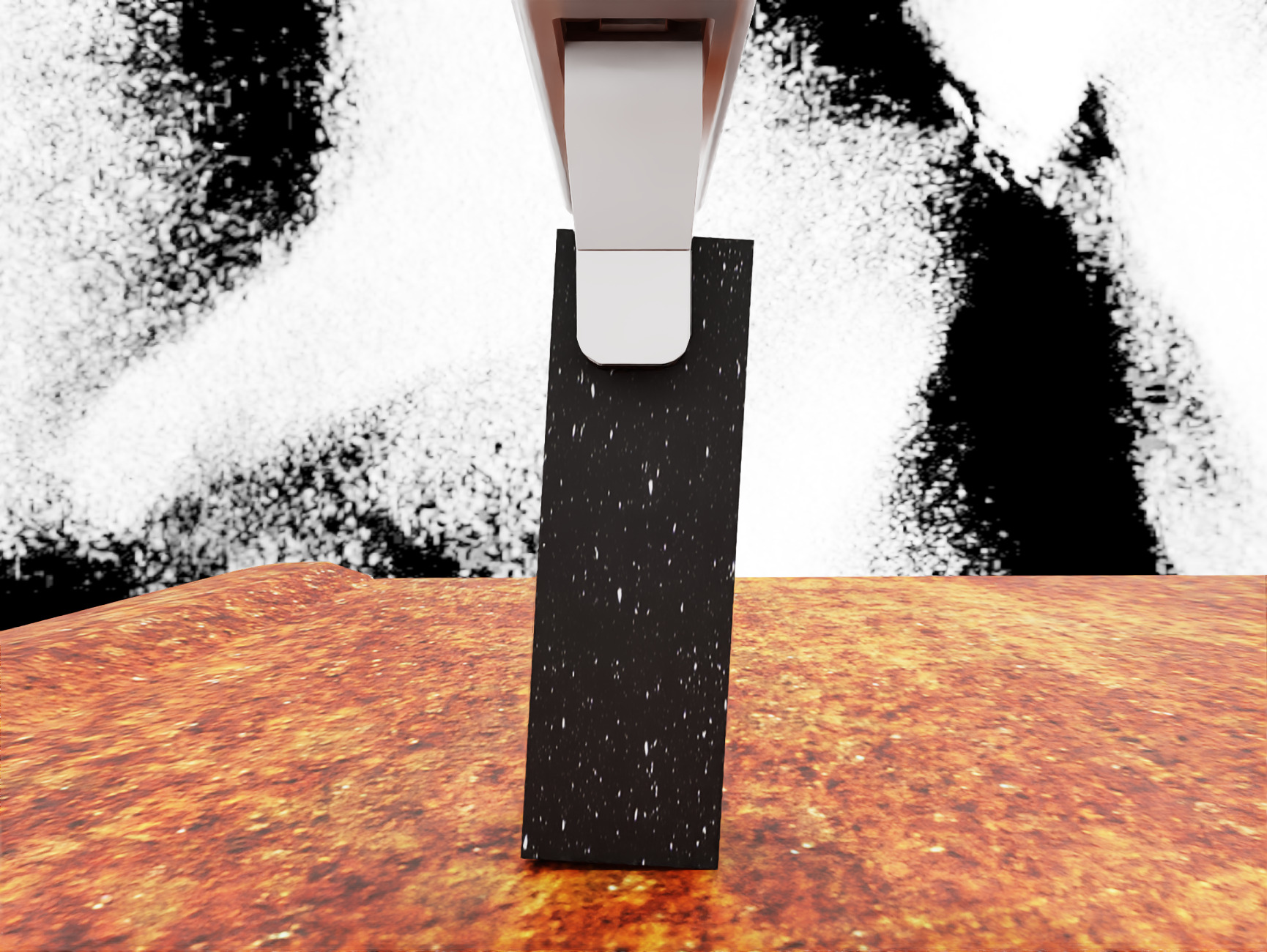}
        \includegraphics[width=\figWidth]{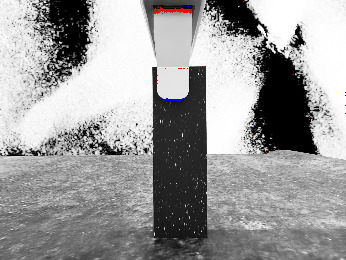}
        \includegraphics[width=\figWidth]{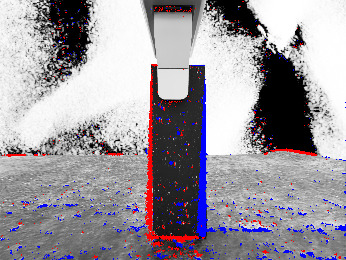}
        \includegraphics[width=\figWidth]{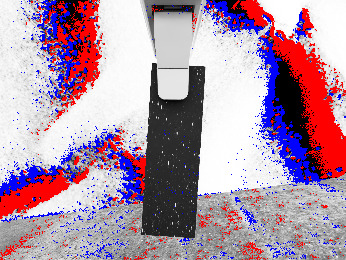}
        \includegraphics[width=\figWidth]{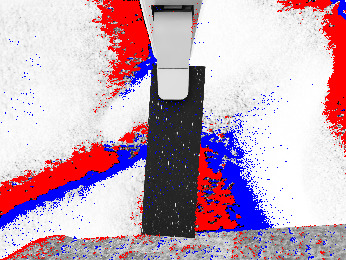}
        \includegraphics[width=\figWidth]{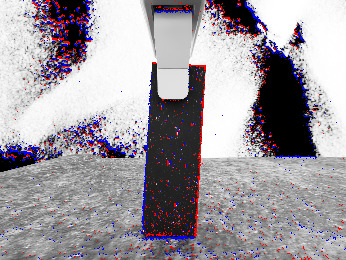}
        \subcaption{Slip sample.}
    \end{subfigure}
    \caption{Visualization of two pick-and-place sequences across five key movement phases (from left to right): (1) initial position, (2) object lifting, (3) lateral tilt, (4) back to vertical gripper orientation, and (5) final placement.
    Top row: color frames from the physics simulator.
    Bottom row: synthesized event data overlaid on grayscale frames.
    \vspace{-2ex}}
    \label{fig:subsample_vis}
\end{figure}

Three datasets were created to validate the data generation pipeline and conduct slip detection experiments: 
two simulated sets (\emph{Simple Set}, \emph{Complex Set}) and a real-world data set (\emph{Real Set}). 
The \emph{Simple Set} consists of 192 simulations using only two texture combinations and little diversity. 
The \emph{Complex Set} contains 1200 simulations incorporating 48 different textures and high diversity, while maintaining the specified parameter constraints. 
This set is split into training and testing subsets. 
These distinct sets serve two purposes: demonstrating the ability to control dataset scale and complexity, and providing varying levels of difficulty for slip detection (\ie, classification) experiments.

Each simulation produces a sample containing approximately 1 million events, 450 colored frames at 1730$\times$1300 pixel resolution, and 2$\times$450 quaternions describing gripper and object orientations, totaling nearly one gigabyte (GB). 
While each pick-and-place task simulation represents 7-8 seconds of virtual time, generating a single sample requires about 6 minutes. 
However, the simulator's parallel processing capabilities allow for simultaneous execution of multiple simulations. 
For the generation of these datasets, 8 simulations were run in parallel, resulting in an average computation time of 2.7 minutes per sample on a system with two NVIDIA RTX A6000 GPUs and a dual AMD EPYC 7543 processor featuring 128 logical cores.

\begin{table}[t]
\centering
\resizebox{\linewidth}{!}{
\setlength{\tabcolsep}{4pt} %
\begin{tabular}{lccc}
\toprule
\multirow{2}{*}{\centering \textbf{Property}} 
& \multirow{2}{*}{\makecell{\textbf{Simple}\\\textbf{Set}}} 
& \multirow{2}{*}{\makecell{\textbf{Complex Set}\\\textbf{Training}}} 
& \multirow{2}{*}{\makecell{\textbf{Complex Set}\\\textbf{Test}}} \\
\\
\midrule
Width of cuboid & 2 & 1000 & 200\\
Height of cuboid & 2 & 1000 & 200 \\
Mass of cuboid & 2 & 1000 & 200 \\
Texture sets & 2 & 1000 & 200 \\
Light sphere position & 2 & 1000 & 200 \\
Light sphere brightness & 1 & 1000 & 200 \\
Background sphere brightness & 1 & 1000 & 200 \\
Gripping position horizontally & 3 & 1000 & 200 \\
Gripping position vertically & 2 & 1000 & 200 \\
\rowcolor[gray]{.9}
Total & 192 & 1000 & 200 \\
\bottomrule
\end{tabular}}
\caption{Selected property values across sets. 
The \emph{Simple Sets} was created by combining the extrema of the selected properties and a continuous background lighting. For both subsets of the \emph{Complex Set}, every parameter value was selected randomly for every simulation, with a strict separation of values between both.\label{tab:properties}
\vspace{-3ex}}
\end{table}

\subsection{Simple Set}
The \emph{Simple Set} consists of 192 samples without an explicit test set and totals 185 GB. 
The data's simplicity is achieved by limiting parameter variations: only two texture combinations for the cuboid, surface, and background, two positions for the sphere light, and extreme values for all other parameters. These extremes include the highest and lowest cuboid masses (to guarantee slip and non-slip cases), maximum and minimum heights, maximum and minimum widths, and gripping positions at the leftmost, middle, and rightmost points, both at the highest and lowest possible positions. All these parameter options were exhaustively combined to generate the samples (see \cref{tab:properties}).
The event simulation parameters, including the contrast sensitivity and upsampling rate, were set to default values to match a DAVIS346 camera \cite{davis346} at an input frame rate of 60 Hz.

\subsection{Complex Set}
\begin{figure}[t]
\def\figHeight{0.18\linewidth}
  \centering
   \begin{subfigure}{\linewidth}
        \centering
        \includegraphics[height=\figHeight]{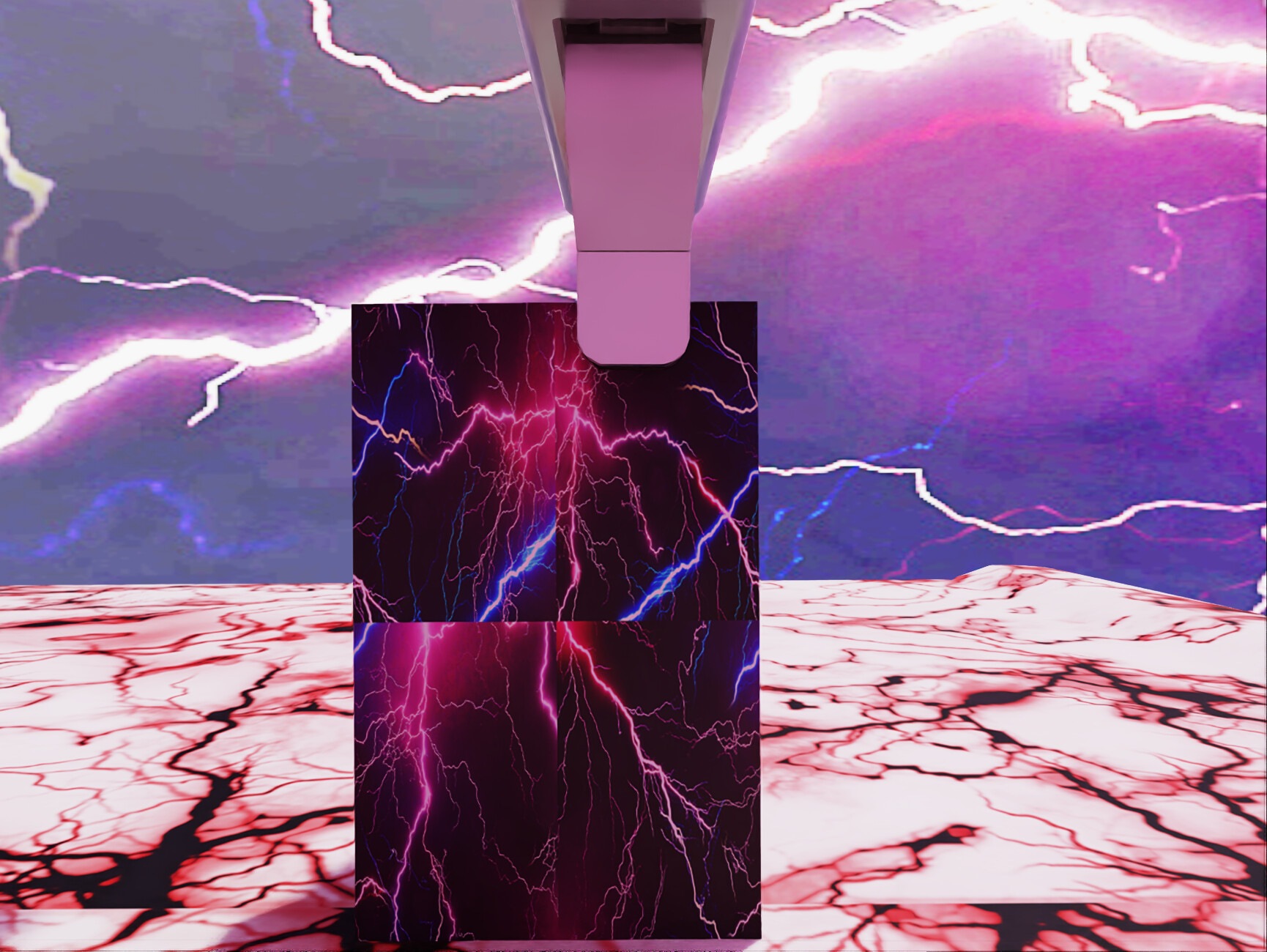}
        \includegraphics[height=\figHeight]{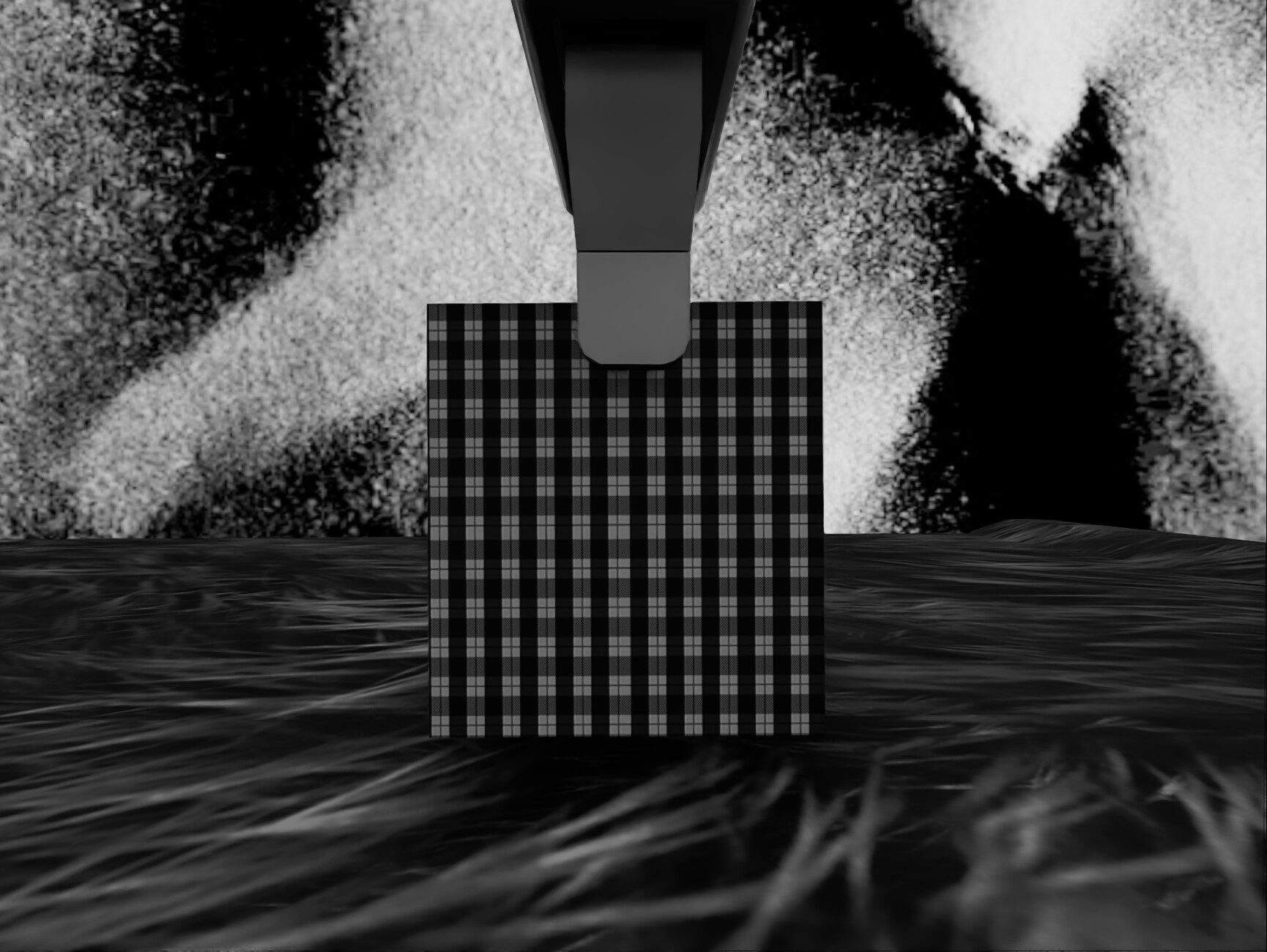}
        \includegraphics[height=\figHeight]{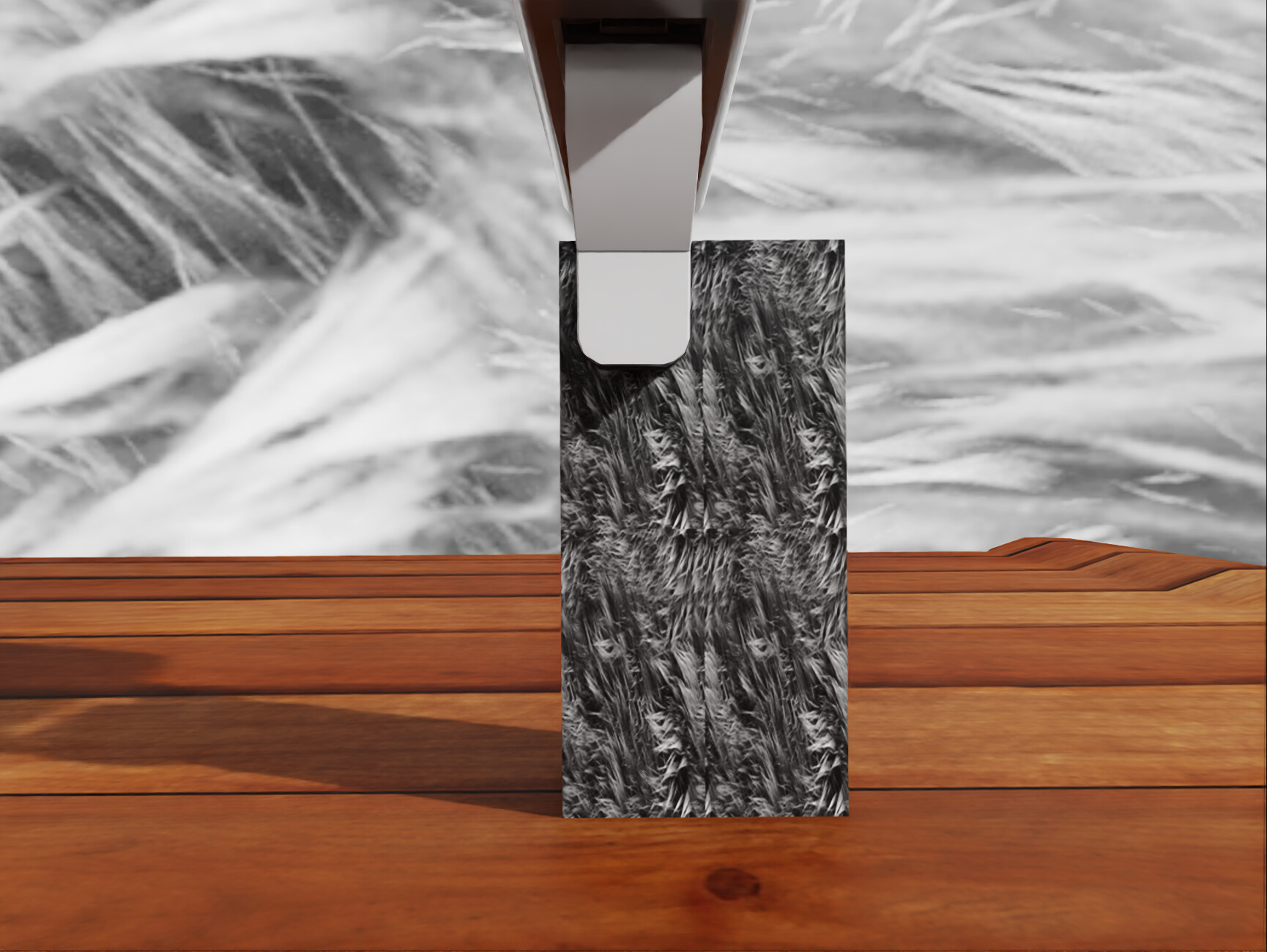}
        \includegraphics[height=\figHeight]{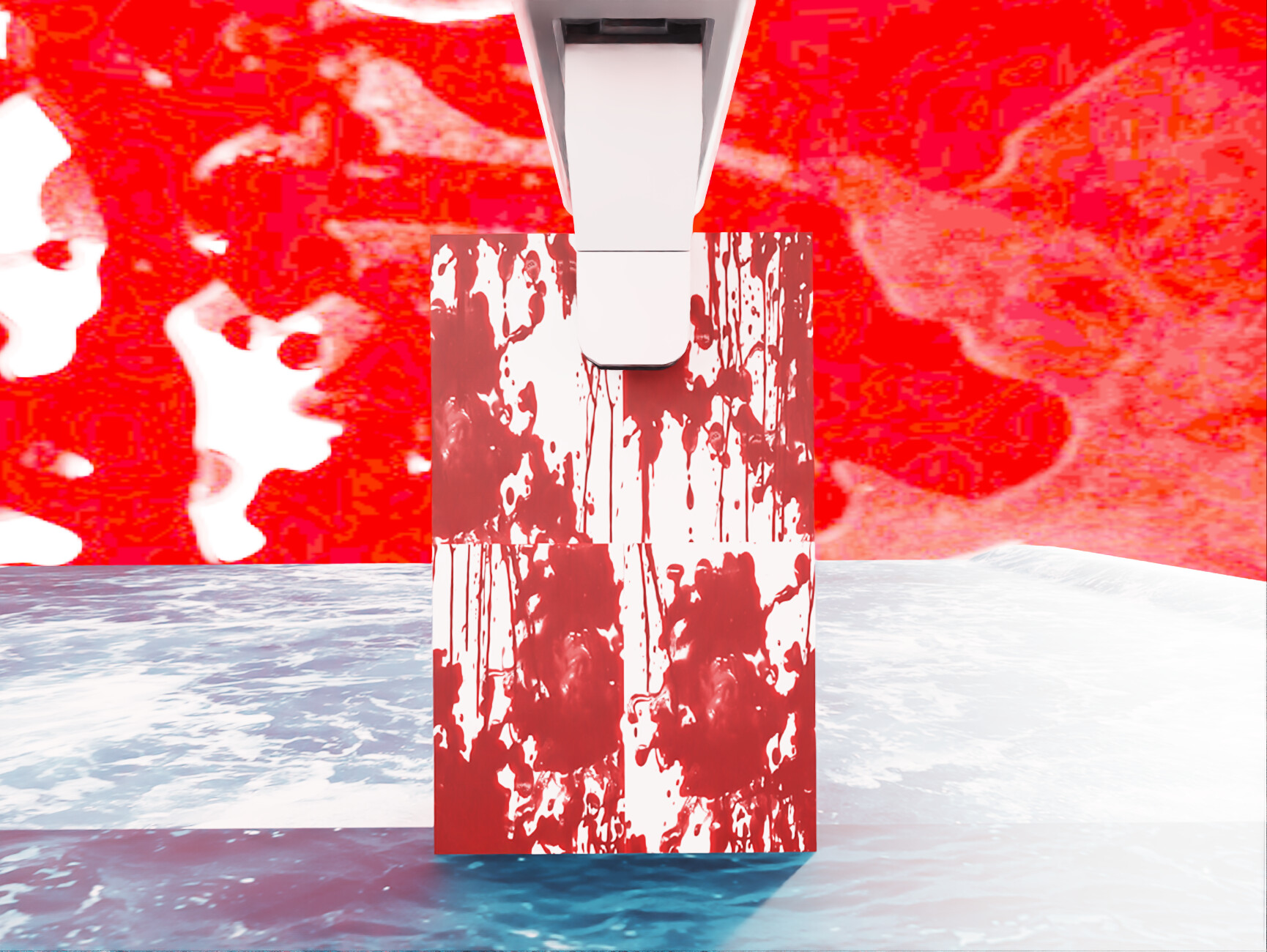}
        \includegraphics[height=\figHeight]{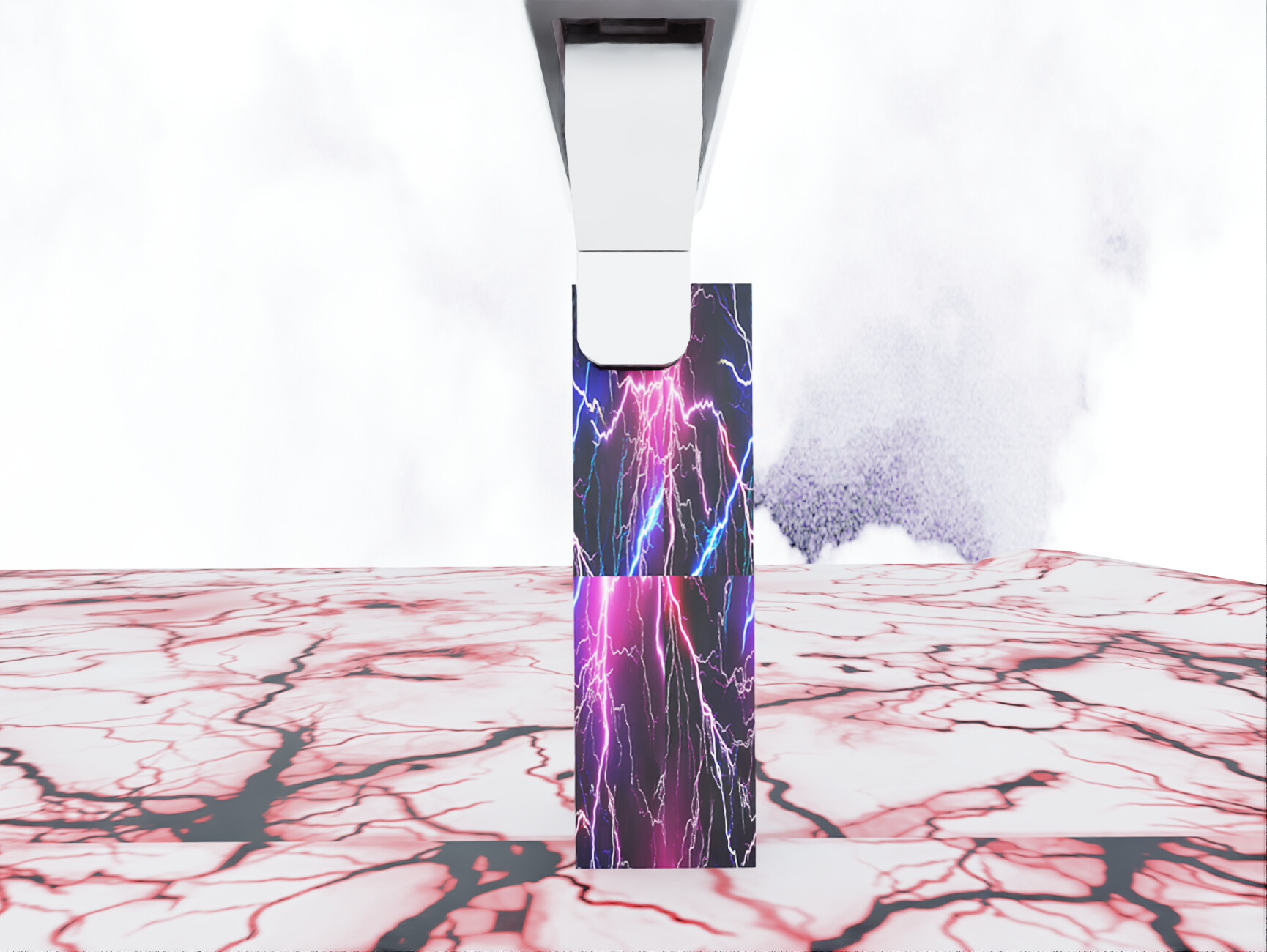}
        \includegraphics[height=\figHeight]{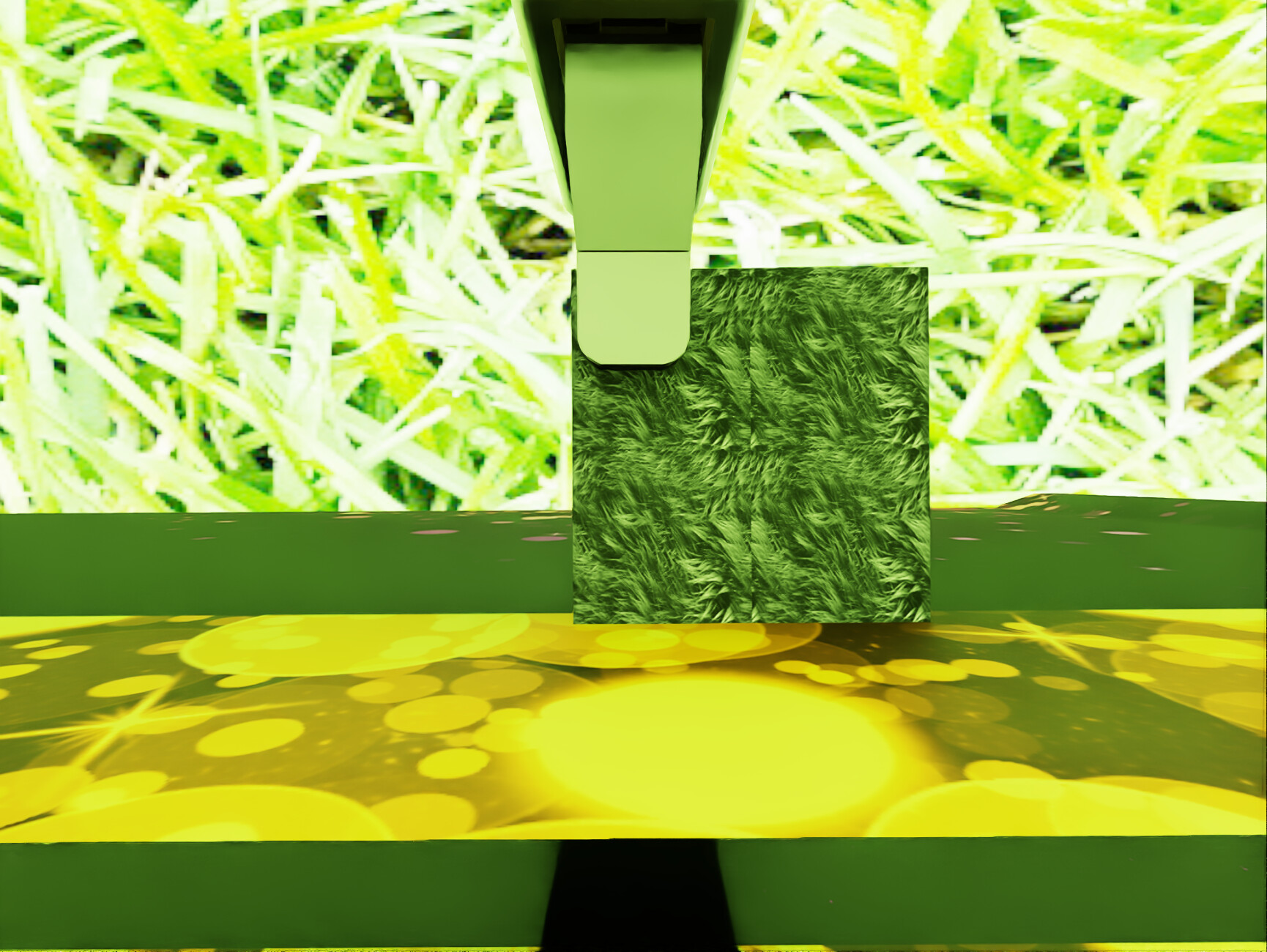}
        \includegraphics[height=\figHeight]{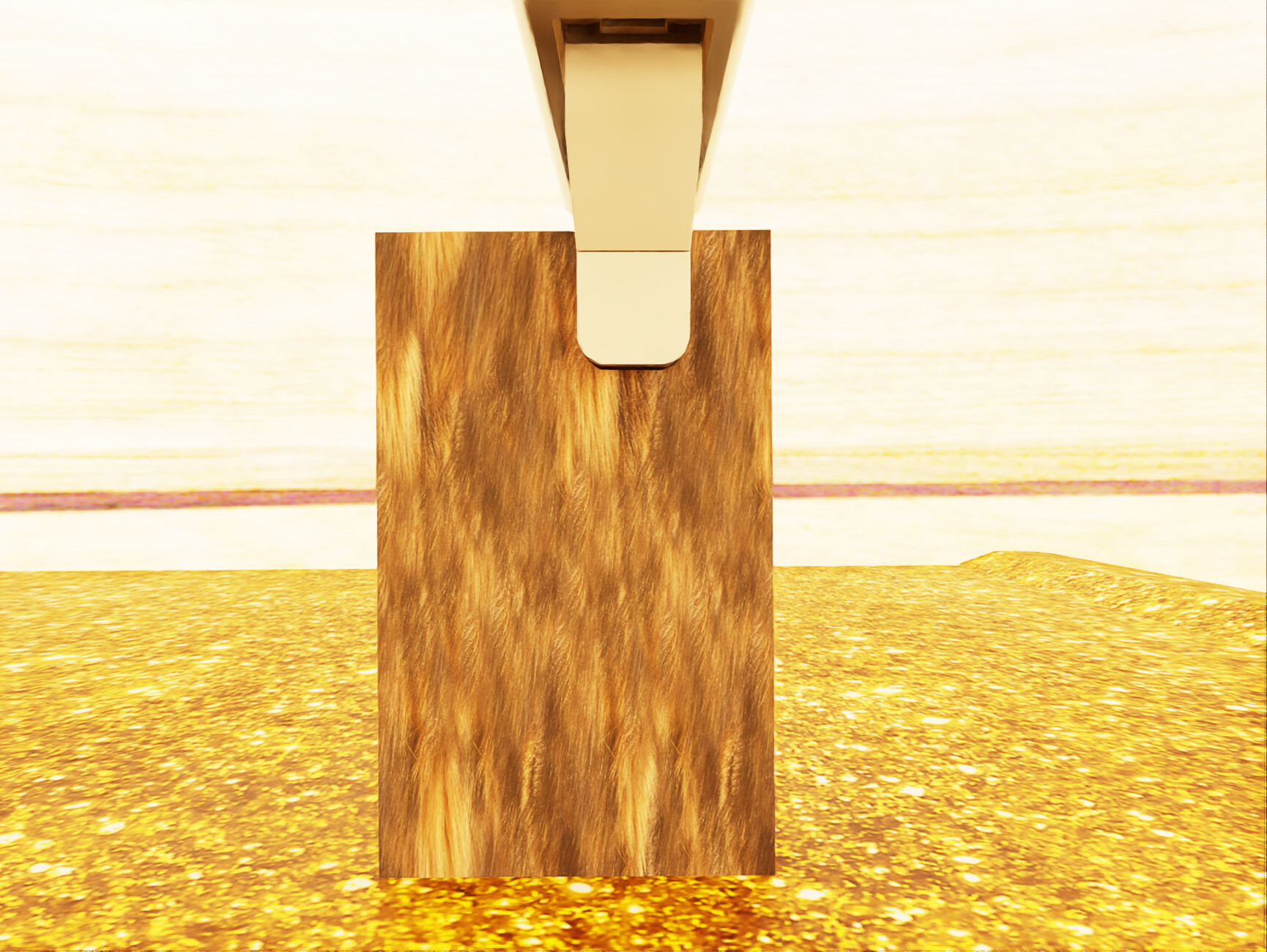}
        \includegraphics[height=\figHeight]{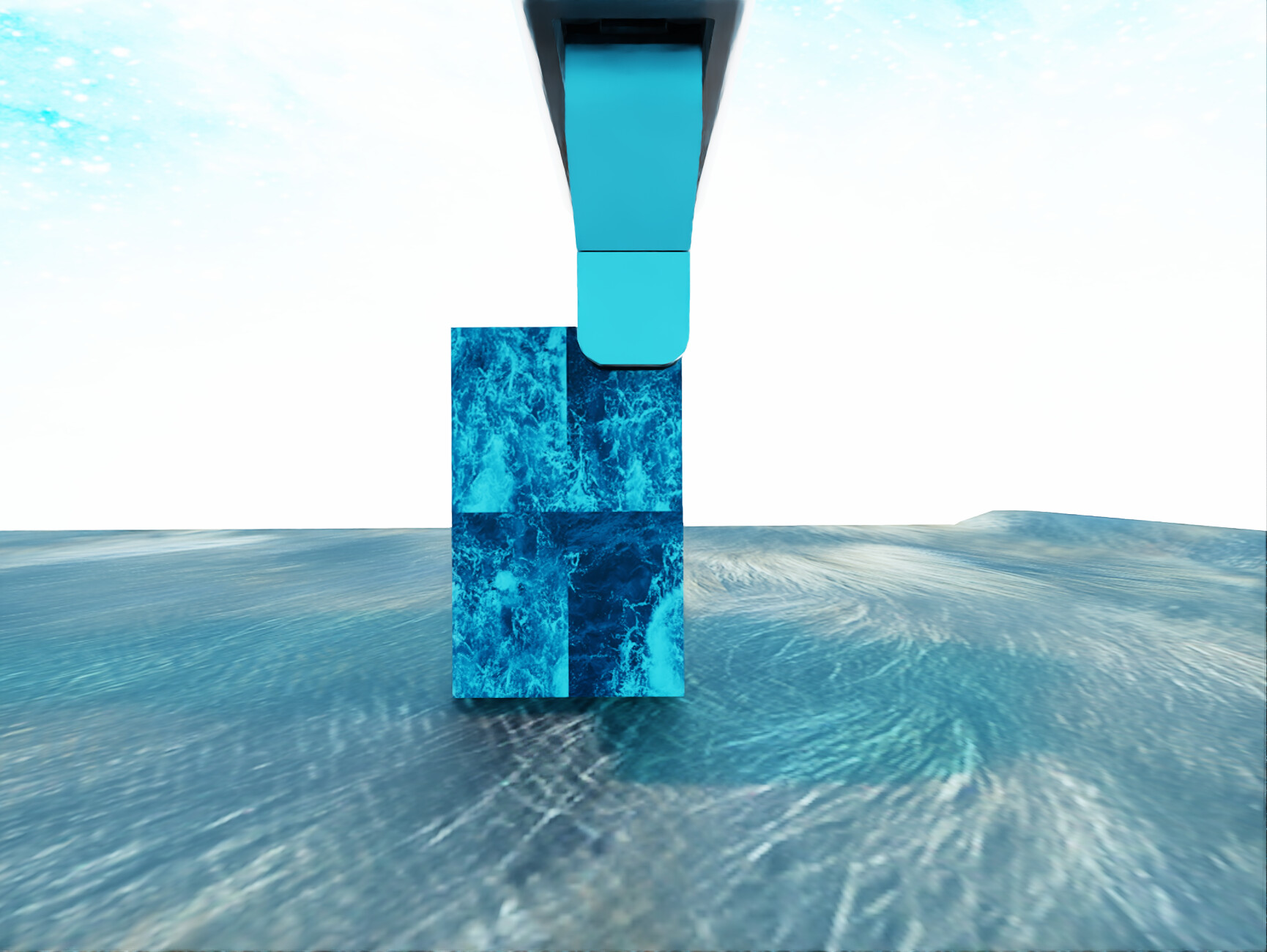}
        \vspace{-1ex}
   \end{subfigure}
   \caption{Frames of the \emph{Complex Set} with varying parameters.\label{fig:random_parameters}\vspace{-1ex}
   }   
\end{figure}

The \emph{Complex Set} (\cref{fig:random_parameters}) consists of a training set of 1000 samples and a test set of 200 samples. 
The training set totals 883 GB, though this size is primarily due to visualization data (images and videos). 
Excluding these visualization files (\cref{fig:subsample_vis}) and retaining only the event files and ground-truth data reduces the storage to approximately 87 GB.
Data diversity was achieved through multiple parameter variations. 
A pool of 48 unique textures was split with an 80:20 ratio between training and test sets, with each sample randomly receiving three textures for the surface, cuboid, and background. 
Illumination varied through two random components: the enclosing sphere's brightness ranged from very bright to dim, while a light sphere was randomly positioned on a hemisphere above the scene, with its intensity varying from bright (casting sharp shadows) to dim (producing minimal shadows) (see \cref{tab:properties}). 
The cuboid's parameters covered shape configurations from wide and flat to tall and thin, with masses ranging from guaranteed slip to guaranteed non-slip conditions. 
The gripper's position was randomized between the object's edges and middle, and from maximum grip depth to the top edge.
While parameter values can occur multiple times within either the training or test set, strict separation between sets was maintained --no exact value appears in both. 
However, due to the continuous nature of the parameters, values can be similar across sets.

The event simulation parameters, including the contrast sensitivity and upsampling rate, were configured to match a DAVIS346 camera \cite{davis346} at an input frame rate of 60 Hz.

\subsection{Real Set}
\begin{figure}[t]
\def\figHeight{0.192\linewidth}
  \centering
   \begin{subfigure}{\linewidth}
        \centering
        \includegraphics[height=\figHeight,angle=180]{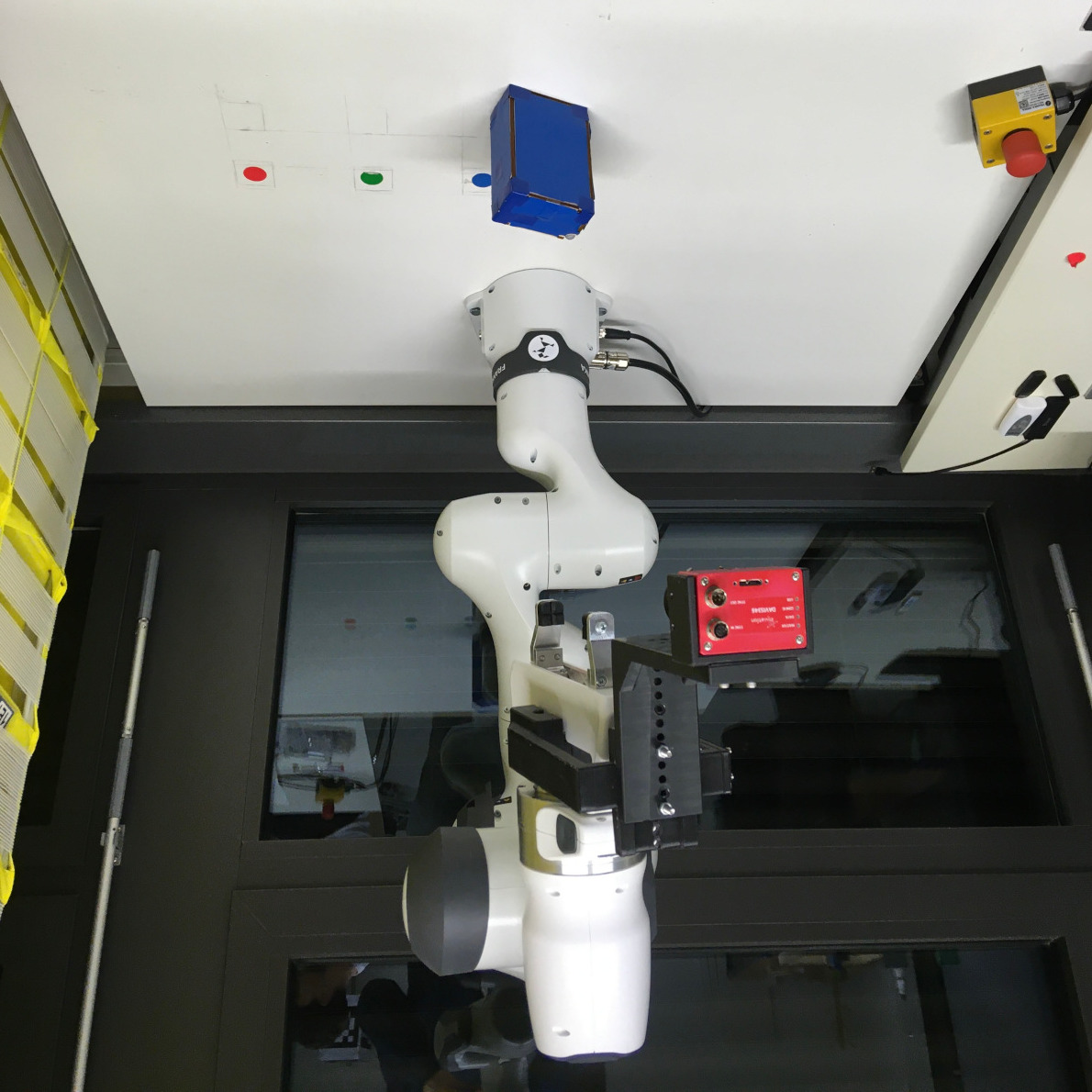}
        \includegraphics[height=\figHeight,angle=180]{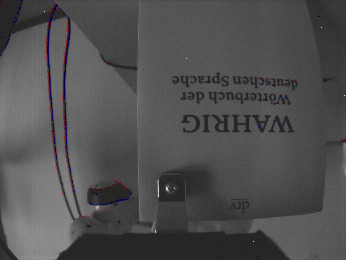}
        \includegraphics[height=\figHeight,angle=180]{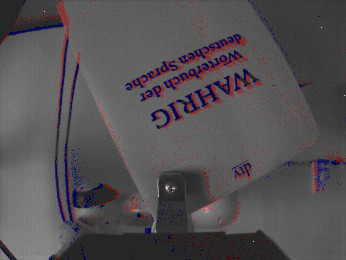}
        \includegraphics[height=\figHeight,angle=180]{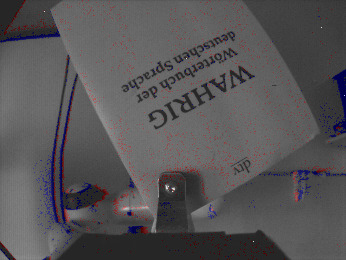}
        \vspace{-1ex}
   \end{subfigure}
   \caption{Left: Experimental setup for real-world data collection. Right: Event data and frames captured during a slip sequence from real-world experiments. \label{fig:real_world_data}\vspace{-2ex}
   }
\end{figure}

The \emph{Real Set} (\cref{fig:real_world_data}) comprises five samples, including three slipping sequences and two non-slipping ones.
The ground-truth labels were assigned through manual observation of the combined frame and event data.
The experimental setup closely follows \cite{albert}, using a Panda robot mounted on a flat white surface with a DAVIS346 camera attached to the gripper. 
The primary modification is a custom 3D-printed camera mount that provides a side view of the gripper and object, replacing the previous front view configuration.
This physical setup served as the reference model for the simulator development.
The manipulation object chosen was a moderately textured book. 
While slightly larger than its simulated counterpart and showing minor deformation during grasping, it is similar to the simulated data.

\section{Experiments}
\label{sec:experiments}
To evaluate the usefulness and validity of the generated datasets, experiments were performed by training different ANN slip-detection architectures. 
The chosen architectures are inspired by the experiments in \cite{VT-SNN} (\cref{sec:neural_networks}). 
The networks include an MLP, an SNN, and a 3-dimensional CNN (3D-CNN).
All three architectures were trained on both sets (\emph{Simple Set} and \emph{Complex Set}), separately, with varying parameters. 
The data was preprocessed and parameter sweeps were performed using \cite{wandb}, with lists of possible values provided for each parameter. 
For every run, parameter values were selected randomly. 
The parameters included batch size, learning rate, and two optimization algorithms: RMSProp \cite{hinton2012neural} 
and Adam \cite{Kingma15iclr}.

\subsection{Preprocessing}
To prepare the data for training and testing, the samples of the \emph{Simple Set}, \emph{Complex Set} and the \emph{Real Set} were preprocessed (\cref{sec:preprocessing}). 
The preprocessing involves splitting the samples into subsamples of 16 ms. These subsamples were then segmented differently for each network architecture: 150 bins of about 1 ms for the SNN and MLP, and 350 bins of about 0.46 ms for the 3D-CNN.
We labeled the subsamples based on their angular difference change using two thresholds (see \cref{sec:ang_diff}): subsamples showing changes greater than $1.0^\circ$ are labeled slip, whereas those with changes less than $0.1^\circ$ are non-slip. 
Subsamples with values between these thresholds were considered uncertain and discarded. To balance the datasets, the larger set of labeled subsamples was reduced by random removal until it matched the size of the smaller set.
Further preprocessing steps included cropping the events to 200$\times$250 px and generating visualizations. Finally, the resulting subsample sets were randomly split into training and validation sets using an 80:20 ratio. 
This resulted in 918 training and 230 validation subsamples for the \emph{Simple Set}, 11554 training, 2888 validation, and 2682 test subsamples for the \emph{Complex Set} and 81 subsamples for the \emph{Real Set}.

\subsection{Slip Detection on Simulated Data}
\label{ref:slip_Detection_Complex}
The results for both the \emph{Simple Set} and \emph{Complex Set} are presented in \cref{tab:results}. The following section focuses on the \emph{Complex Set} results, while a detailed description of the \emph{Simple Set} is provided in the Supplementary Material.

\begin{table}[t]
\centering
{\small
\setlength{\tabcolsep}{2.5pt} %
\begin{tabular}{lcccc}
\toprule
\multirow{2}{*}{\centering \textbf{Architecture}} & 
\multirow{2}{*}{\centering \textbf{Trained on}} 
& \multirow{2}{*}{\makecell{\textbf{Validation}\\\textbf{Accuracy}}} 
& \multirow{2}{*}{\makecell{\textbf{Test}\\\textbf{Accuracy}}} 
& \multirow{2}{*}{\makecell{\textbf{Real Data}\\\textbf{Accuracy}}} \\
\\
\midrule
MLP & \emph{Simple Set} & 98\% & N/A & N/A \\
SNN & \emph{Simple Set} & 96\% & N/A & N/A \\
3D-CNN & \emph{Simple Set} & 99\% & N/A & N/A \\
MLP & \emph{Complex Set} & 72\% & 69\% & 63\%\\
SNN & \emph{Complex Set} & 70\% & 69\% & 59\%\\
3D-CNN & \emph{Complex Set} & 80\% & 78\% & 63\%\\
\bottomrule
\end{tabular}
}
\caption{Data-driven slip detection performance across datasets. Classification accuracies evaluated on the \emph{Simple Set} (validation only), the \emph{Complex Set} (validation and test), and the \emph{Real Set} (pre-trained on \emph{Complex Set}).
\label{tab:results}
\vspace{-3ex}
}
\end{table}

We also evaluated on the \emph{Complex Set} using the same three ANNs (MLP, SNN, and 3D-CNN), comprising 225 training runs in total.
They achieved lower accuracies than on the Simple Set, with the 3D-CNN achieving the highest validation accuracy at 80\%, while both SNN and MLP performed at approximately 70\%. 
Importantly, all three architectures demonstrated consistent performance when evaluated on the test set, with test accuracies closely matching their respective validation accuracies.

For the MLP architecture, the parameter sweep yielded its best performance with a validation accuracy of 72\% and a cross entropy loss of 0.001. This was achieved using the Adam optimizer with a learning rate of 0.0005 and a batch size of 512.
Most validation accuracies ranged between 62\% and 72\%, with a few runs resulting in 50\% accuracy. The experiments revealed that lower learning rates (0.005 and below) consistently outperformed higher values, while batch sizes between 128 and 1024 yielded superior accuracies compared to smaller batch sizes below 128.
Final evaluation on the \emph{Complex Set} test set demonstrated an accuracy of 69\% with a corresponding loss of 0.009.

For the SNN architecture, the parameter sweep yielded its best performance with a validation accuracy of 70\% and a spike loss of 1175. This was achieved using the RMSProp optimizer with a learning rate of 0.00005 and a batch size of 512.
Most runs achieved validation accuracies between 67\% and 69\%, with a few runs resulting in accuracies below 55\%. The higher performing configurations consistently used learning rates between 0.00005 and 0.006, and generally performed better with batch sizes above 256.
Final evaluation on the \emph{Complex Set} test set demonstrated an accuracy of 69\% with a corresponding loss of 1262.

For the 3D-CNN, the parameter sweep achieved its best performance with 80\% accuracy and 0.026 cross entropy loss, using the RMSProp optimizer, a learning rate of 0.0001, and a batch size of 16.
High-performing runs ranging from 75\% to 80\% accuracy, moderate runs around 70\%, and poor performers between 50\% to 60\%. 
Consistent with previous findings, good performance correlated with smaller batch sizes (16 to 256) and learning rates (0.0001 to 0.002).
The trained network was also evaluated on the test set, achieving an accuracy of 78\% and a loss of 0.01.

\subsection{Slip Detection on Real Set}
The three best-performing ANNs from \Cref{ref:slip_Detection_Complex} were evaluated on the \emph{Real Set}.
The networks achieved accuracies of 63\% (MLP, cross entropy loss: 0.06), 59\% (SNN, spike loss: 1336), and 63\% (3DCNN, cross entropy loss: 0.8) on this dataset (see \cref{tab:results}), which indicates the models' capability to process real-world data.

\section{Conclusion}
\label{sec:conclusion}

This work presents a simulator for studying slip detection with event cameras in pick-and-place tasks, leveraging the event-camera's ability to isolate relative motion. 
It supports creating large datasets with configurable complexity, enabling both analytical and data-driven approaches. 
It features photorealistic rendering, physics-based simulation, synthetic event generation, and rich visualization for dataset analysis and edge-case investigation.

Two simulated datasets were created and used to validate the simulator’s effectiveness for slip detection applications through both qualitative visual analysis and machine learning evaluation.
The visual analysis involved assessing the photorealistic quality of images, the realism of object movements, and the ability of the human eye to differentiate between slips and non-slips in the images and the visualized events. 
The results were positive: the images exhibited a high degree of photorealism, object movements appeared natural, and most slip events were visually detectable.

All three ANNs (MLP, SNN and 3D-CNN) achieved validation accuracy rates above 95\% on the \emph{Simple Set}.
While training on the \emph{Complex Set} yielded lower accuracies, the performance on a never-before-seen test set containing new textures and scenarios remained consistent with the validation results. The 3D-CNN demonstrated the strongest performance, achieving 80\% validation accuracy and 78\% accuracy on the test set.
These consistent results between validation and test sets suggest the models generalized well to new data rather than overfitting to the training set.
Initial evaluation on a small real-world dataset showed accuracies around 60\%, indicating the potential for slip detection transfer from simulation to real-world applications.

\section*{Acknowledgment}
Funded by the Deutsche Forschungsgemeinschaft (DFG, German Research Foundation) under Germany’s Excellence Strategy – EXC 2002/1 ``Science of Intelligence'' – project number 390523135.
Funded by Amazon Research Award.
We thank Profs.~Toussaint and Raisch for support with the robotic manipulator and its software. 

\ifarxiv
\clearpage
\section*{Supplementary Material}
\subsection*{Slip Detection on Simple Set}
A total of 495 training runs were executed over all three neural networks (MLP, SNN, and 3D-CNN). 
Each ANN achieved validation accuracy exceeding 95\% with optimal parameter settings, with the 3D-CNN reaching 99.59\% accuracy. 
The results are reported in \cref{tab:results}.

The MLP parameter sweep achieved its best performance with 98\% validation accuracy and 0.003 cross-entropy loss, using the RMSProp optimizer, a learning rate of 0.003, and a batch size of 128. 
Accuracies showed distinct clustering around either 90\% or 50\%. Poor results tended to occur with combinations of smaller batch sizes and higher learning rates. RMSProp and Adam optimizers showed comparable performance. For high-accuracy runs, loss values ranged from 0.0007 to 0.1.

The SNN's best configuration achieved 96\% validation accuracy with a spike loss of 178, using the RMSProp optimizer, a learning rate of 0.011, and a batch size of 512. 
Training showed robust performance with accuracies consistently near or above 90\% across all parameter combinations. Loss values varied widely, with some runs around 280 and others below 60 or even 20.

The 3D-CNN achieved its peak performance with 99.59\% validation accuracy and 0.0035 cross-entropy loss, using the RMSProp optimizer, a learning rate of 0.002, and a batch size of 16. Accuracies clustered either above 80-90\% or around 50\%. High accuracies were predominantly achieved with small batch sizes (8 to 32) and learning rates below 0.020. Loss values stayed under 0.045, with high-accuracy runs typically below 0.01.

\fi 

{\small
\bibliographystyle{ieee_fullname}
\bibliography{all,egbib}
}

\ifarxiv
\else
\clearpage
\section*{Supplementary Material}
\subsection*{Slip Detection on Simple Set}
A total of 495 training runs were executed over all three neural networks (MLP, SNN, and 3D-CNN). 
Each ANN achieved validation accuracy exceeding 95\% with optimal parameter settings, with the 3D-CNN reaching 99.59\% accuracy. 
The results are reported in \cref{tab:results}.

The MLP parameter sweep achieved its best performance with 98\% validation accuracy and 0.003 cross-entropy loss, using the RMSProp optimizer, a learning rate of 0.003, and a batch size of 128. 
Accuracies showed distinct clustering around either 90\% or 50\%. Poor results tended to occur with combinations of smaller batch sizes and higher learning rates. RMSProp and Adam optimizers showed comparable performance. For high-accuracy runs, loss values ranged from 0.0007 to 0.1.

The SNN's best configuration achieved 96\% validation accuracy with a spike loss of 178, using the RMSProp optimizer, a learning rate of 0.011, and a batch size of 512. 
Training showed robust performance with accuracies consistently near or above 90\% across all parameter combinations. Loss values varied widely, with some runs around 280 and others below 60 or even 20.

The 3D-CNN achieved its peak performance with 99.59\% validation accuracy and 0.0035 cross-entropy loss, using the RMSProp optimizer, a learning rate of 0.002, and a batch size of 16. Accuracies clustered either above 80-90\% or around 50\%. High accuracies were predominantly achieved with small batch sizes (8 to 32) and learning rates below 0.020. Loss values stayed under 0.045, with high-accuracy runs typically below 0.01.

\fi

\end{document}